\documentclass[10pt,journal,compsoc]{IEEEtran}

\ifCLASSOPTIONcompsoc
  \usepackage[nocompress]{cite}
\else
  \usepackage{cite}
\fi
\usepackage[neverdecrease]{paralist}
\usepackage{multirow}

\ifCLASSINFOpdf
  \usepackage[pdftex]{graphicx}
\usepackage{subfig}
\else
\fi
\usepackage{tikz}
\usetikzlibrary{bayesnet}
\usepackage{epstopdf}

\usepackage[cmex10]{amsmath}
\usepackage{amssymb}

\usepackage{algorithmic}
\usepackage{algorithm}

\usepackage{enumitem}

\usepackage{url}
\newcommand{\N}{\mathcal{N}}

\newcommand{\bs}{\boldsymbol}
\newcommand{\bx}{\textbf{x}}
\newcommand{\bff}{\textbf{f}}
\newcommand{\by}{\textbf{y}}
\newcommand{\bz}{\textbf{z}}
\newcommand{\bw}{\textbf{w}}
\newcommand{\bX}{\textbf{X}}

\newcommand{\GP}{\mathcal{G}\mathcal{P}}
\newcommand{\transpose}{{\mbox{\scriptsize T}}}

\begin{document}
%
\title{A Bayesian additive model for understanding public transport usage in special events}
%
%
%
%

\author{Filipe~Rodrigues,
        Stanislav~S.~Borysov,
        Bernardete~Ribeiro,~\IEEEmembership{Senior~Member,~IEEE},
        and~Francisco~C.~Pereira,~\IEEEmembership{Member,~IEEE}
\IEEEcompsocitemizethanks{\IEEEcompsocthanksitem F.~Rodrigues is with the Technical University of Denmark (DTU), Bygning 115, 2800 Kgs. Lyngby, Denmark. E-mail: rodr@dtu.dk\protect
\IEEEcompsocthanksitem S.~S.~Borysov is with the Singapore--MIT Alliance for Research and Technology, 1 CREATE Way, 138602 Singapore and Nordita, KTH Royal Institute of Technology and Stockholm University, Roslagstullsbacken 23, SE-106 91 Stockholm, Sweden.
\IEEEcompsocthanksitem B.~Ribeiro is with the CISUC, Department of Informatics Engineering, University of Coimbra, 3030-290 Coimbra, Portugal.
\IEEEcompsocthanksitem F.~C.~Pereira is with the Massachusetts Institute of Technology (MIT),
77 Massachusetts Avenue, 02139 Cambridge, MA, USA and Technical University of Denmark (DTU), Bygning 115, 2800 Kgs. Lyngby, Denmark.}
\thanks{10.1109/TPAMI.2016.2635136 \mbox{https://ieeexplore.ieee.org/document/7765036/}}}

%
%

\markboth{IEEE Transactions on Pattern Analysis and Machine Intelligence,~Vol.~39, No.~11, 2113-2126
}%
{A Bayesian additive model for understanding public transport usage in special events}
%



\definecolor{cool_blue}{RGB}{48,127,175}
\definecolor{cool_red}{RGB}{225,101,89}

\IEEEtitleabstractindextext{%
\begin{abstract}
Public special events, like sports games, concerts and festivals are well known to create disruptions in transportation systems, often catching the operators by surprise. Although these are usually planned well in advance, their impact is difficult to predict, even when organisers and transportation operators coordinate. The problem highly increases when several events happen concurrently. To solve these problems, costly processes, heavily reliant on manual search and personal experience, are usual practice in large cities like Singapore, London or Tokyo. 
This paper presents a Bayesian additive model with Gaussian process components that combines smart card records from public transport with context information about events that is continuously mined from the Web. We develop an efficient approximate inference algorithm using expectation propagation, which allows us to predict the total number of public transportation trips to the special event areas, thereby contributing to a more adaptive transportation system. Furthermore, for multiple concurrent event scenarios, the proposed algorithm is able to disaggregate gross trip counts into their most likely components related to specific events and routine behavior. Using real data from Singapore, we show that the presented model outperforms the best baseline model by up to 26\% in $R^2$ and also has explanatory power for its individual components. 
\end{abstract}

}

\maketitle

\IEEEdisplaynontitleabstractindextext

%
\IEEEpeerreviewmaketitle

\ifCLASSOPTIONcompsoc
\IEEEraisesectionheading{\section{Introduction}\label{sec:introduction}}
\else
\section{Introduction}
\label{sec:introduction}
\fi

%
%
%
%
\IEEEPARstart{F}{or} environmental and societal reasons, public transport has a key role in the future of our cities. However, the challenge of tuning public transport supply adequately to the demand is known to be complicated. While typical planning approaches rely on understanding habitual behavior \cite{krygsman2004}, it is often found that our cities are too dynamic and difficult to predict. A particularly disruptive case is with special events, like concerts, sports games or festivals \cite{kwon2006}. In practice, public transport operators easily understand and prepare for the very large ones (e.g. Formula~1 races, football finals, Olympic games), but find it hard to tune for smaller ones, particularly when multiple events co-occur. Partly, the problem so far has been due to lack of appropriate data: operators and planners could hardly know the full list of events, and trip counting has been generally incomplete and inefficiently acquired. 

On the other hand, even with a complete dataset, understanding and predicting impacts of future events would not be a humanly simple task, as there are many dimensions involved. One needs to consider details such as the type of a public event, popularity of the event protagonists, size of the venue, price, time of day and still account for routine demand behavior, as well as the effect of other co-occurring events. In other words, besides data, sound computational methodologies are also necessary to solve this multidimensional problem. The generalization of smart card technologies in public transport systems together with the ubiquitous reference to such events on the Internet, which provides abundance of social information, effectively proposes a potential solution to these limitations \cite{pereira2013JITS}. 
However, the question of modeling transportation demand in special event scenarios with an efficient and accurate model has remained an unmet challenge. 

Under the assumption of decomposition of public transport demand into routine commuting and special events contributions, a natural response to this problem lies in the realm of additive models. These models allow one to deal with a variable number of components, which is particularly suitable for addressing special events scenarios. In this paper, we propose to combine this concept with the Bayesian framework to model transportation behavior in a tractable manner by exploiting the recent advances in model-based machine learning \cite{Bishop2013} and approximate inference \cite{Minka2001}. With this aim, a Bayesian additive model is proposed, where each of a variable number of additive components is modeled as a Gaussian process (GP) \cite{Rasmussen2005} with truncated Gaussian or Poisson outputs, whose values are then summed up to produce the observed totals. 

\begin{figure}[!t]
\centering
\includegraphics[scale=.33,trim = 10mm 37mm 10mm 5mm]{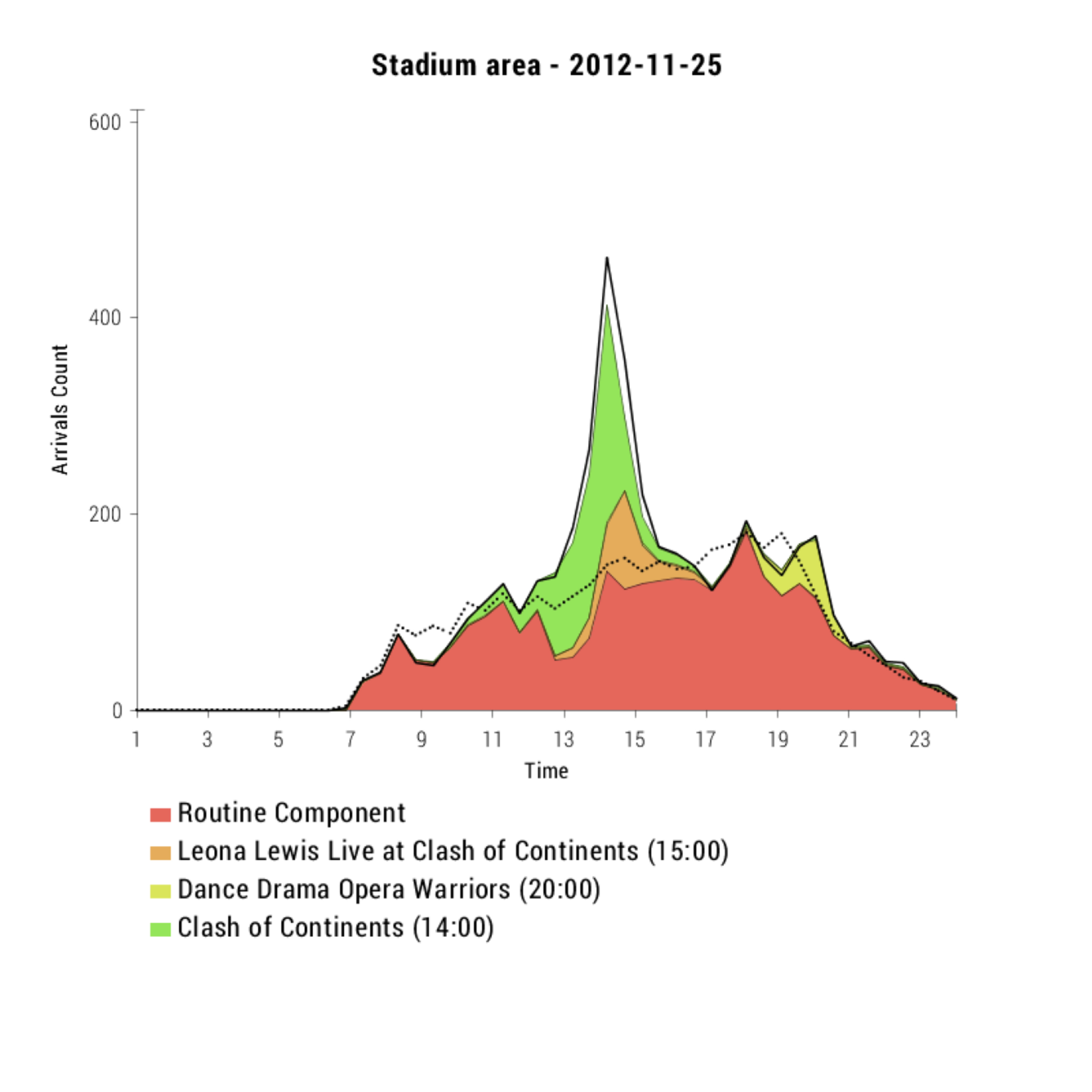}
\caption{Breakdown of the observed time-series of subway arrivals (black solid line) into the routine commuting (area in red) and the contributions of events (orange, yellow and green areas). The dotted line represents the median arrivals over all the days in the observed data that correspond the same weekday. Events start times are shown in parentheses.}
\label{fig:example}
\end{figure}

Besides making predictions of the number of public transport trip arrivals in a given place, by employing such a general-purpose Bayesian additive framework, it is possible to breakdown an observed time-series of arrivals into the contributions of the different components: routine commuting and individual special events. For example, Fig.~\ref{fig:example} illustrates this application with actual results from the proposed model using real data from the Singapore's Indoor Stadium and Kallang Theatre, which share the same subway stop. On this day (November 25, 2012), the Indoor Stadium had a tennis tournament (``Clash of Continents'') and the Kallang Theatre had an opera. Within the tennis tournament, there was a short Leona Lewis concert scheduled between two exhibition matches, sometime between 15:00 and 16:00. Besides the records from the public transport service, the proposed model uses multiple features from the Web that will be described later, including start and end times, venue, results from an automated web search, etc. In this example, it identifies that the large bulk of trips between 12:00 and 15:00 were arrivals to the Leona Lewis concert and to the tennis tournament. Then, after 17:00, there were arrivals to the opera (scheduled for 20:00) together with routine trips.  

By applying the proposed framework to the problem of modeling public transport arrivals under special events scenarios, we are therefore able to (i) predict the distribution of the total number of arrivals that will be observed in the future considering all the events that are spatially and temporally close; (ii) disaggregate the time-series of arrivals into the contributions of a ``routine" component (e.g. commuting) and a variable number of event components, making predictions about the contribution of each future event separately. All this information can be of great value not only for public transport operators and planners, but also for event organizers and public transport users in general. Finally, by using a Bayesian approach, the proposed framework can be easily adapted to perform online learning. 
Together with the efficient approximate inference algorithm developed, the proposed model has the ability to scale to very large datasets and to be deployed in practice. 

Although we focus on a transportation application, it is important to note that this is a general-purpose methodology that can be extended to different application domains, such as electrical signal disaggregation or source separation. 
Indeed, the Bayesian additive framework described in this paper can be of great value for any prediction task where knowing the importance (or contribution) of different inputs is required. 
For example, when modeling air quality through the presence of pollutants, it is essential to have interpretable systems, so that researchers can understand how each individual factor (traffic, forest fires, kitchens, air conditioning/heating, industry, etc.) contributes to the total values observed or forecasted. 
The same applies to transport management challenges, where operators and planners need to understand what originates demand fluctuations to mitigate them properly.

Finally, with this paper we hope to encourage researchers from other domains to adopt model-based machine learning approaches \cite{Bishop2013}, thereby designing solutions that incorporate their domain knowledge and that are specifically tailored for their problems, instead of forcing their data into standard machine learning algorithms. 

The remainder of this paper is organized as follows. In the next section, we review the relevant literature for this work. Section~\ref{section:problem_fromulation} provides motivation for the proposed Bayesian additive model (BAM), while the model itself is introduced and explained in Section~\ref{section:BAM}. The corresponding experimental results are presented in Section~\ref{section:Experiments}. The paper ends with the conclusions.

\section{Literature review}
\label{sec:literature}

\subsection{Transport demand modeling for special events}
From the perspective of transportation research and practice, the main methodology to demand modeling is provided by the discrete choice framework \cite{ben-akiva-lerman-1987}, which combines logistic and probit regression machinery with behavior modeling and utility theory. Two recent examples of this approach include Kuppam et al \cite{Kuppam2011} and Chang and Lu \cite{chang2013}, who propose a 4-step model approach, where they predict, for each event, the number of trips by type, trip time-of-day, trip origin/destination (OD), mode and vehicle miles travelled/transit boardings generated due to the events. Although these works seem behaviorally sound and provide plenty of detail, they are highly dependent on survey response data and, in fact, do not explicitly consider event characteristics. For example, trip ODs have been demonstrated to vary by event type \cite{CalabresePervasive2010}. 

Particularly for mega-events (e.g. Olympic games, World cup, Formula 1), research works and best practices are available (e.g. \cite{FHWA06,CoutroubasEtAl03}). However, as Potier et al \cite{PotierEtAl03} point out, even for such events, transport demand is more difficult to forecast than habitual mobility, particularly in the case of open-gate events. In facing these constraints, authorities tend to rely on trial and error experience (for recurring events), checklists (e.g. \cite{FHWA06}) and sometimes invest in a reactive approach rather than planning, as happens in Germany, with the Real-time Traffic and Traveller Information (RTTI) and its active traffic management \cite{Bolte2006}, and in the Netherlands \cite{Middleham2006}. However, such tools have limited applicability, particularly for smaller and medium-sized events, which are harder to capture and evaluate. 

In order to cover a wider spectrum of events, one needs to access a more comprehensive source of information about social activities. Previous studies have shown that information contained on announcement websites, social networks and other online sources is indeed of practical value for transportation demand prediction models \cite{pereira2013JITS,pereira2015so}. However, the inclusion of this information in such models is not trivial and still remains an open research problem. 
An initial approach was made in \cite{pereira2015so} to explain the amount of excessive demand (in a continuous period, called \emph{hotspot}) with an additive model implemented in Infer.NET \cite{InferNET12}. However it relied on linear regression components, which is a very limitative assumption. Furthermore, since the authors focus only on explaining hotspots, they did not account for a routine component as we do in the proposed model. In this paper, we approach the more difficult and more general problem of public transport arrivals prediction, a problem with higher and more direct practical implications.

\subsection{Additive models}

Linear regression models provide an effective and attractively simple framework for understanding how each input variable relates with the observed target variables. However, they fail to capture non-linear dependencies between inputs and target variables, which are recurrent in the real-world. On the other hand, flexible models such as neural networks or Gaussian processes (GPs) lay on the opposite side of the spectrum, where the target variables are modeled as complex non-linear functions of all input variables simultaneously. Unfortunately, due to their black-box nature, the interpretativeness and the ability to understand how each input is contributing to the observed target are typically lost. Additive models \cite{Hastie1990} contrast with these by specifying the target variable to be the result of a linear combination of non-linear  functions of the individual inputs. Due to this structured form, additive models provide an interesting tradeoff between interpretability and flexibility. 

The typical approach in additive models is to rely on scatterplot smoothers for representing non-linear effects of the individual inputs in a flexible way \cite{Hastie2003,Ravikumar2009}. Additive models can then easily be fitted using a backfitting procedure \cite{Hastie1990}, which iteratively fits each of the scatterplot smoothers to the residuals of the sum of the estimates of all the other smoothers, until a convergence criterion is met. The model proposed in this paper contrasts with these works in several ways: (i) instead of simple scatterplot smoothers we consider the use of Gaussian processes with different output distributions such as truncated Gaussian or Poisson; and (ii) we propose a fully Bayesian approach for inferring the posterior distribution of the individual function values using expectation propagation \cite{Minka2001}.

From the specific perspective of Gaussian processes, Duvenaud et al. \cite{Duvenaud2011} proposed the additive GP: a GP model for functions that decompose into a sum of other low-dimensional functions. This is achieved through the development of a tractable kernel which allows additive interactions of all orders, ranging from univariate functions of the inputs to multivariate interactions between all input variables simultaneously. Despite being efficient in exploring all orders of interaction between inputs, additive GPs do not support a variable number of interacting functions as we require for our practical application of public transport demand prediction, where there is a variable number of events happening. Furthermore, the Bayesian additive framework presented in this paper is more flexible, in the sense that it allows to incorporate further restrictions on the models such as non-negativity constraints, as well as combining linear with non-linear functions or combining GPs with different covariance functions. 

Compared to the ensemble learning literature, the proposed models shares several characteristics with Bayesian additive regression trees (BART) \cite{Chipman2010}. Namely, both approaches model the observations as a sum of non-linear functions. However, in BART these functions are tree-based weak learners. Hence, contrarily to the model proposed in this paper, BART is not designed for generating interpretable models.

\subsection{Data fusion}

The proposed Bayesian additive model can also be seen from the perspective of data fusion in a urban computing scenario \cite{zheng2015methodologies}, since it combines data from two heterogeneous sources: smartcard data on transportation usage and event data mined from the Internet. Urban data fusion approaches are often divided into three main categories \cite{zheng2014urban}: (i) approaches that treat different data sources equally and put together the features extracted from various data sources into one single feature vector, (ii) approaches that use different sources of data at different stages and (iii) approaches that feed different datasets into different parts of a model simultaneously. The model proposed in this paper belongs to the third category, which is recognized to be the most challenging one, but also the one that typically provides the best results \cite{zheng2013u,zheng2014urban}. Indeed, our experimental results provide further evidence that fusing different data sources at model level is better than fusing data at feature level. To the best of our knowledge, the proposed Bayesian additive model is the first to explicitly model and quantify the effect of events in transport demand using a large-scale data-driven approach. 

\section{Problem formulation}\label{section:problem_fromulation}

Let $y$ be the total number of public transport arrivals at a given time. The most natural approach to model $y$ is to consider it to be a function of time, the day of the week, whether or not it is a holiday, etc. We refer to these as routine features, $\bx^r$, as they characterize the routine behavior of a given place. A wide majority of previous works focuses solely on these features (\mbox{e.g.} \cite{Van2015}). However, as previously discussed, there are several other dynamic aspects of transportation demand that need to be accounted for. Particularly, we are interested in the effect of special events. Let $\bx^{e_i}$ be a feature vector characterizing a given event $e_i$, such as the venue, categories, tags, etc. Since the number of events that occur in a given area varies, we consider models of the form
\begin{align}
y = f_r(\bx^r) + \sum_{i=1}^E f_e(\bx^{e_i}) + \epsilon,
\label{eq:additive_formulation}
\end{align}
where $\epsilon \sim \N(0,v)$ is the observation noise and $E$ denotes the number of events that can affect the observed arrivals $y$. Hence, the number of events, $E$, varies between observations. However, if we assume the functions $f_r(\bx^r)$ and $f_e(\bx^{e_i})$ to be linear functions of their inputs, parameterized by a vector of coefficients $\bw_r$ and $\bw_e$ respectively, then we can write (\ref{eq:additive_formulation}) as
\begin{align}
y &= (\bw_r)^\transpose \bx^r + (\bw_e)^\transpose \Bigg( \sum_{i=1}^E \bx^{e_i} \Bigg) + \epsilon = \bw^\transpose \bx + \epsilon,
\label{eq:lr_simplification}
\end{align}
where we defined $\bw \triangleq [\bw_r; \bw_e]$ and $\bx \triangleq [\bx^r; \sum_{i=1}^E \bx^{e_i}]$. 

As we can see, in the case of linear functions, the feature vectors of all events can be aggregated by summation, which reduces the problem to a simple linear regression. However, that does not allow us to properly explore our domain knowledge. For example, we might want to assign the routine and event components specific distributions (\mbox{e.g.} Poisson or truncated Gaussian) in order to impose certain properties on their values, such as non-negativity. Furthermore, for the particular application domain of transportation demand, the functions $f_r(\bx^r)$ and $f_e(\bx^{e_i})$ can be non-linear, as our experiments demonstrate (see Section~\ref{section:Experiments}). As soon as we start considering non-linear models, the distributive law used in (\ref{eq:lr_simplification}) can no longer be applied, and the feature vectors of the different events $\bx^{e_i}$ cannot be simply summed up. Moreover, the number of such vectors varies constantly from time to time. Aggregating these in order to allow a standard regression formulation, such as Gaussian process regression, to be applied is then a non-trivial research question. In the following section, we propose a Bayesian additive framework that not only allows us to handle these issues but also leads to other attractive properties.

\section{Bayesian additive model} 
\label{section:BAM}

\subsection{Model description} 

The proposed Bayesian additive model builds on the assumption that there is a base routine component $y^r = f_r(\bx^r)$ and a variable number of event components $y^{e_i} = f_e(\bx^{e_i})$, whose contributions are summed up to obtain the total observed arrivals $y$ in a given area. Since we wish to constrain the values of the individual components, $y^r$ and $\{y^{e_i}\}_{i=1}^E$, to be non-negative, we define the latter to be one-side truncated Gaussians, which we denote as
\begin{align}
\label{eq:truncated_routine}
y^r &\sim \mathbb{I}(y^r > 0) \, \N(y^r | f_r(\bx^r), \beta_r),\\
\label{eq:truncated_events}
y^{e_i} &\sim \mathbb{I}(y^{e_i} > 0) \, \N(y^{e_i} | f_e(\bx^{e_i}), \beta_e), 
\end{align}
where $\mathbb{I}(a > 0)$ is an indicator function that takes the value 1 if and only if $a > 0$, $\beta_r$ and $\beta_e$ are the variances of routine and events components, respectively. 
Alternatively, we also consider a variant of the model that assumes the component values to be Poisson distributed with an exponential link function, such that
\begin{align}
\label{eq:poisson_routine}
y^r &\sim \mbox{Poisson} (y^r | e^{f_r(\bx^r)}),\\
\label{eq:poisson_events}
y^{e_i} &\sim \mbox{Poisson} (y^{e_i} | e^{f_e(\bx^{e_i})}). 
\end{align}
In either case, the observed totals $y$ are then defined as the sum of all components
\begin{align}
y = y^r + \textstyle\sum_{i=1}^E y^{e_i} + \epsilon, \quad \epsilon \sim \N(0,v).
\end{align}
For some applications, such as energy disaggregation, we might wish to add constraints to the totals as well \cite{zhong2014signal}.

Having specified the additive structure of the model, the next step is to specify how to model the functions $f_r$ and $f_e$. Perhaps the simplest approach would be to assume $f_r(\bx^r)$ and $f_e(\bx^{e_i})$ to be linear functions of their inputs as in (\ref{eq:lr_simplification}). However, as we shall see in Section~\ref{section:Experiments}, for the transport application considered in this paper, these can be highly non-linear. We therefore propose the use of Gaussian processes (GPs) \cite{Rasmussen2005} to model these functions. GPs are flexible non-parametric Bayesian models that fit well within the probabilistic modeling framework \cite{Barber2012} and are state-of-the-art methods for many classification and regression tasks. In the transportation field, GPs have shown to achieve state-of-the-art results for various tasks, such as travel-time prediction \cite{ide2009travel}, traffic volume forecasting \cite{xie2010gaussian} and public transit flows prediction in urban areas \cite{neumann2009stacked}. Although in this paper we focus on Gaussian processes, it is important to note that the additive framework described above is general enough to allow a large variety of models to be applied. 

Letting the vectors $\bff^r$ and $\bff^e$ denote the functions $f_r(\bx^r)$ and $f_e(\bx^{e_i})$ evaluated for all feature vectors $\bx^r$ and $\bx^{e_i}$ respectively, Gaussian process modeling proceeds by placing a GP prior on $\bff^r$ and $\bff^e$, such that $\bff^r \sim \GP(m_r(\bx^r) \equiv 0, k_r(\bx^r,\bx^{r^\prime}))$ and $\bff^e \sim \GP(m_e(\bx^{e}) \equiv 0, k_e(\bx^{e},\bx^{e^\prime}))$, where for the sake of simplicity (and without loss of generality, having our data centered) we assumed the GPs to have zero mean so that the GPs are completely defined in terms of the covariance functions $k_r$ and $k_e$. One key advantage of the proposed additive framework is then the ability to assign different covariance functions to the various components. This could be useful in many applications like, for example, for modeling seasonal trends by using periodic covariance functions. However, for the particular application considered in this paper, we shall use squared exponential covariance functions of the form
\begin{align}
\label{eq:covariance_function}
k(\bx,\bx^\prime) = \sigma_f^2 \exp\Big(- \frac{1}{2} (\bx - \bx^\prime)^\transpose \textbf{M} \, (\bx - \bx^\prime)\Big),
\end{align}
where $\textbf{M} = \mbox{diag}(\bs\ell)^{-2}$ and $\bs\ell = \{\ell_1,...,\ell_D\}$ are the characteristic length-scales. Since the inverse of the length-scales determines how relevant each input dimension is, this covariance function can be used for automatic relevance determination (ARD) by finding the hyper-parameter values $\bs\ell$ that maximize the marginal likelihood of the data.

\begin{figure}[t!]
\begin{center}
\includegraphics{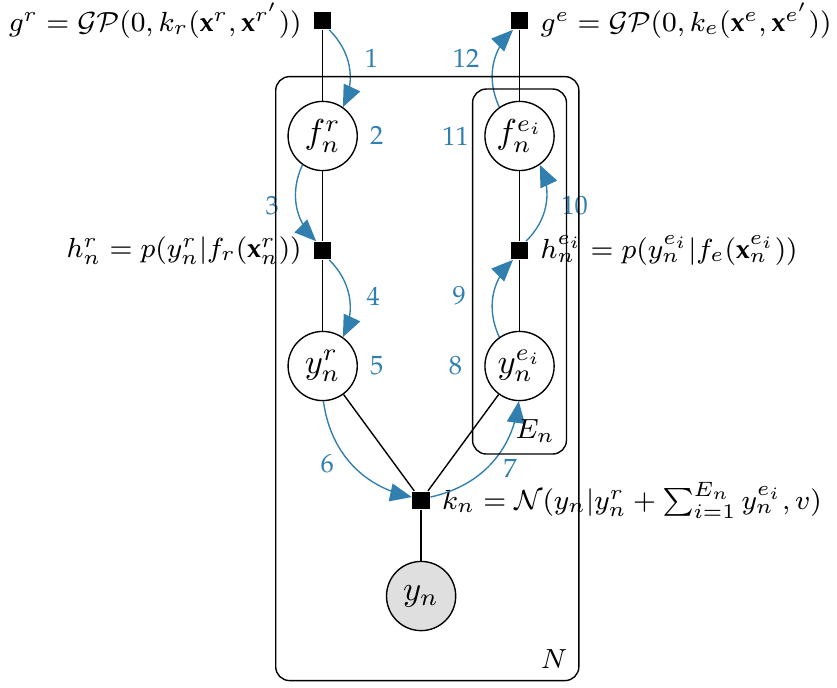}
\caption{Factor graph of the proposed Bayesian additive model with Gaussian process components. 
The blue arrows represent the message-passing algorithm for performing approximate Bayesian inference. The second flow of messages starting from the GP factor for the events component that goes in the opposite direction is not shown.}
\label{fig:factor_graph_gp}
\end{center}
\end{figure}

The generative process of the proposed Bayesian additive model can then be summarized as follows:
\begin{enumerate}
\item Draw $\bff^r \sim \GP(0, k_r(\bx^r,\bx^{r^\prime}))$
\item Draw $\bff^e \sim \GP(0, k_e(\bx^e,\bx^{e^\prime}))$
\item For each observation $n \in \{1,...,N\}$
\begin{enumerate}
\item Draw routine component $y_n^r \sim p(y_n^r | f_r(\bx_n^r))$
\item For each event $e_i$, $i \in \{1,...,E_n\}$
\begin{enumerate}
\item Draw event contribution $y_n^{e_i} \sim p(y_n^{e_i} | f_e(\bx_n^{e_i}))$
\end{enumerate}
\item Draw total observed arrivals\\$y_n \sim \N(y_n | y_n^r + \sum_{i=1}^{E_n} y_n^{e_i}, v)$
\end{enumerate}
\end{enumerate}
\vspace*{0.3cm}
where $E_n$ denotes the number of events that are associated with the $n^{th}$ observation. The conditional distributions $p(y_n^r | f_r(\bx_n^r))$ and $p(y_n^{e_i} | f_e(\bx_n^{e_i}))$ are either the truncated Gaussians from (\ref{eq:truncated_routine}) and (\ref{eq:truncated_events}) or the Poissons from (\ref{eq:poisson_routine}) and (\ref{eq:poisson_events}). In practice, the choice of these conditional distributions is purely problem-specific and the proposed model and its inference algorithm are flexible to allow for a wide variety of alternatives. Fig.~\ref{fig:factor_graph_gp} shows a factor graph representation of the proposed model, which will be particularly useful in the following section for deriving a message passing algorithm to perform approximate Bayesian inference using expectation propagation (EP) \cite{Minka2001}. 

\subsection{Approximate inference}
\label{sec:approximatE_nnference}

Let $\mathcal{D}$ be a dataset of $N$ observations, each one corresponding to the total number of arrivals (transportation demand) associated with the bus or subway stations that serve a certain special events area in a given time interval. Formally, $\mathcal{D} = \{ \bx_n^r, \bX_n^e, y_n \}_{n=1}^N$, with $\bX_n^e = \{ \bx_n^{e_i} \}_{i=1}^{E_n}$, and $\bx_n^r$, $\bx_n^{e_i}$ being the attributes of the routine and events components of observation $n$, respectively.

Given a dataset $\mathcal{D}$, our goal is two-fold: (i) compute the marginal distributions of the individual components $y_n^r$ and $y_n^{e_i}$ and (ii) make predictions for new input vectors $\{\bx_*^r, \bX_*^e\}$. According to the factor graph in Fig.~\ref{fig:factor_graph_gp}, the joint distribution of the proposed model with truncated Gaussian components is given by
\begin{align}
\label{eq:joint_prob}
&p(\bff^r,\bff^e,\by^r,\textbf{Y}^e,\by|\{\bx_n^r,\bX_n^e\}_{n=1}^N)\\ 
&= \N(\bff^r|\textbf{0},\textbf{K}^r) \, \N(\bff^e|\textbf{0},\textbf{K}^e) \prod_{n=1}^N \mathbb{I}(y_n^r > 0) \, \N(y_n^r|f_n^r,\beta_r) \nonumber\\
&\times \Bigg( \prod_{i=1}^{E_n} \mathbb{I}(y_n^{e_i} > 0) \, \N(y_n^{e_i}|f_n^{e_i},\beta_e) \Bigg) \, \N\Bigg(y_n \Bigg| y_n^r + \sum_{i=1}^{E_n} y_n^{e_i}, v\Bigg),\nonumber
\end{align}
where we defined $\by \triangleq \{y_n\}_{n=1}^N$, $\by^r \triangleq \{y_n^r\}_{n=1}^N$ and $\textbf{Y}^e \triangleq \{\by_n^e\}_{n=1}^N$, with $\by_n^e \triangleq \{y_n^{e_i}\}_{i=1}^{E_n}$. The covariance matrices $\textbf{K}^r$ and $\textbf{K}^e$ are obtained by evaluating the covariance functions $k_r(\bx^r,\bx^{r^\prime})$ and $k_r(\bx^e,\bx^{e^\prime})$ respectively between every pair of inputs. The joint distribution for the variant with Poisson components is obtained simply by replacing the truncated Gaussians with the Poissons in (\ref{eq:poisson_routine}) and (\ref{eq:poisson_events}).

Unfortunately, the non-Gaussian terms (truncated Gaussian or Poisson) deem exact Bayesian inference computationally intractable. Hence, we proceed by developing a message-passing algorithm using expectation propagation (EP) in order to perform approximate Bayesian inference in the proposed model. In EP, the marginals $p(y_n^r)$ and $p(y_n^{e_i})$ are approximated via moment matching, 
thus resulting in the Gaussian distributions $q(y_n^r)$ and $q(y_n^{e_i})$ with the same mean and variances as $p(y_n^r)$ and $p(y_n^{e_i})$.
EP is then able to approximate the non-Gaussian factors by local Gaussian approximations. However, this approximation is made in the context of all the remaining factors, which gives EP the ability to make approximations that are more accurate in regions of high posterior probability \cite{Minka2001}. 

A particularly useful way of applying EP to larger graphical models is by viewing EP as message-passing in a factor graph. Let the message sent from factor $f$ to variable $x$ be $m_{f \rightarrow x}(x)$. Similarly, let $m_{x \rightarrow f}(x)$ be the message sent from variable $x$ to factor $f$. We can obtain a message-passing viewpoint of EP by defining the following update equations \cite{Murphy2012}
\begin{align}
m_{f \rightarrow x}(x) &= \int f(x,\bz) \prod_{z \in \bz} m_{z \rightarrow f}(z) \, d\bz,\\
\label{eq:m_x_f}
m_{x \rightarrow f}(x) &= \prod_{h \in H_x \backslash \{f\}} m_{h \rightarrow x}(x) = \frac{q(x)}{m_{f \rightarrow x}(x)},\\
q(x) &= \mbox{proj}\Bigg[\prod_{f \in F_x} m_{f \rightarrow x}(x)\Bigg],
\end{align}
where $h \in H_x \backslash \{f\}$ is used to denote all factors $h$ in the neighborhood of $x$, $H_x$, except the factor $f$, $F_x$ denotes the set of factors in the neighborhood of $x$, and the projection operation $\mbox{proj}[p(x)]$ corresponds to finding the approximate distribution $q(x)$ that matches the moments of $p(x)$. 

Making use of this viewpoint of EP, we derive a message-passing algorithm for performing approximate Bayesian inference in the proposed model which consists of 12 steps, as illustrated in Fig.~\ref{fig:factor_graph_gp}, that are iterated until convergence. 
Despite not represented in Fig.~\ref{fig:factor_graph_gp}, there is a second flow of messages starting from the GP factor for the event components that goes in the opposite direction of the one depicted. In practice, these two message flows in opposite directions are implemented in parallel for aditional efficiency. Also, as the figure suggests, all the messages correspond to 1-dimensional Gaussians, which are initialized to be uniform, \mbox{i.e.} zero-mean and infinite variance. In order to reduce oscillation and improve the convergence of the EP algorithm, a dampening procedure is applied, in which the parameters of the messages are linear combinations between the updated message and the message from the previous iteration.

A detailed description of the message-passing algorithm for the proposed model is provided in Appendix. Perhaps the main difficulty in deriving a message-passing algorithm for the proposed model with truncated Gaussian components is in computing the moments of a truncated Gaussian, which can be evaluated analytically as
\begin{align}
\label{eq:moments_truncated}
\mathbb{I}(x > 0) \, \N(x|\mu, \sigma^2) &\approx \hat{Z} \, \N(x|m_x,v_x)\\
m_x &= \mu + \sigma \frac{\N(z)}{\Phi(z)}\nonumber\\
v_x &= \sigma^2 \Bigg(1 - z \frac{\N(z)}{\Phi(z)} -  \bigg(\frac{\N(z)}{\Phi(z)} \bigg)^2 \Bigg)\nonumber\\
\hat{Z} &= \Phi(z),\nonumber
\end{align}
where $z \triangleq \mu/\sigma$ and $\Phi$ is the Gaussian cumulative distribution function. A derivation of these moments is provided in the supplementary material.\footnote{\url{http://www.fprodrigues.com/supp-mat-bam.pdf}} As for the Poisson variant of the proposed model, we use Gauss-Hermite quadrature in order to compute the required integrals over the Poisson. 

The proposed framework provides general applicability and extensibility, which are two significant advantages. In fact, the building blocks for many interesting extensions have already been laid down through the proposed model. For instance, one could extend the model to account for effects of weather or seasonality in the observed arrivals. This could simply be done by including seasonality features in the routine component, but we could go a step further and introduce a new separate GP component, as this would allow us to estimate the effect of seasonality in the observed transportation demand. In fact, the equations for the new messages would be similar to the ones for the routine component, although in this case we might not constrain the marginals to take only non-negative values. Similarly, we could consider including additional components to model higher-level iterations between events, although that would significantly increase the computational complexity of the inference algorithm and make the decomposition results harder to analyse. Therefore, there is a wide variety of interesting applications that could be developed just by making small adaptations to the proposed model and its inference algorithm in Appendix.

\subsection{Marginal likelihood}
\label{sec:marg_lik}

In Bayesian inference, it is often of interest to compute the marginal likelihood of the data $p(\mathcal{D})$. This could be useful, for example, for model comparison. In the case of the proposed Bayesian additive model with Gaussian processes, the marginal likelihood can be used for setting the values of the hyper-parameters of the model by type-II maximum likelihood. Particularly, optimizing the marginal likelihood \mbox{w.r.t.} the length-scales $\bs\ell$ of the ARD covariance function allows us to find the most relevant features. 

The marginal likelihood of the proposed model can be obtained by integrating over the latent variables, giving
\begin{align}
p(\by|\{\bx_n^r&,\bX_n^e\}_{n=1}^N)\nonumber\\
&= \int p(\bff^r,\bff^e,\by^r,\textbf{Y}^e,\by|\{\bx_n^r,\bX_n^e\}_{n=1}^N) \, d\bff^r \, d\bff^e \, d\by^r \, d\textbf{Y}^e, \nonumber
\end{align}
which is intractable due to the non-Gaussian factors $p(y_n^r | f_r(\bx_n^r))$ and $p(y_n^{e_i} | f_e(\bx_n^{e_i}))$ in (\ref{eq:joint_prob}). Luckily, at convergence, the EP algorithm provides us with local approximations to these factors in the form of two unnormalized Gaussian messages which we can use in place of the exact factors while keeping track of the normalization constants, thus making the integral analytically tractable. 

\subsection{Predictions}
\label{sec:predictions}

In the Section~\ref{sec:approximatE_nnference}, we discussed how to compute the posterior distribution of the latent variables given the observed totals using EP. This allows us to understand how transportation demand breaks down as a sum of a routine component and the contributions of the various events that take place in the neighborhood of a given bus or subway station. This, by itself, is of great value for public transport operators, urban planners and event organizers. However, we also want to make predictions for the ``shares" of upcoming events and, ultimately, for the total estimated demand.

Let $\bx_*^r$ be the features of the routine component for a given time and date, and let $\bX_*^e = \{\bx_n^{e_i}\}_{i=1}^{E_n}$ be the set of feature vectors characterizing the events that will take in that place. The EP algorithm in Appendix provides us with approximate posterior distributions for $\bff^r$ and $\bff^e$ given by $q(\bff^r) = \N(\bff^r|\bs\mu^r,\bs\Sigma^r)$ and $q(\bff^e) = \N(\bff^e|\bs\mu^e,\bs\Sigma^e)$. 
These estimates can be used to compute the predictive mean and variance of $f_*^r$ and $\{f_*^{e_i}\}_{i=1}^{E_*}$, as in standard Gaussian process regression and classification. The predictive mean and variance for $f_*^r$ are then given by \cite{Rasmussen2005}
\begin{align}
\mathbb{E}_q[f_*^r|\bff^r,\bx_*^r,\{\bx_n^r\}_{n=1}^N] &= (\textbf{k}_*^r)^\transpose (\textbf{K}^r + \tilde{\bs\Sigma}^r)^{-1} \tilde{\bs\mu^r} \nonumber\\
\mathbb{V}_q [f_*^r|\bff^r,\bx_*^r,\{\bx_n^r\}_{n=1}^N] &= k_r(\bx_*^r,\bx_*^r) - (\textbf{k}_*^r)^\transpose (\textbf{K}^r + \tilde{\bs\Sigma}^r)^{-1} \textbf{k}_*^r ,\nonumber
\end{align}
and similarly for the events variables $\{f_*^{e_i}\}_{i=1}^{E_*}$. We can then use the predictive mean and variance for $f_*^r$ to estimate the share of the routine component as
\begin{align}
p(y_*^r|&\bff^r,\bx_*^r,\{\bx_n^r\}_{n=1}^N) \nonumber\\
&= \mathbb{I}(y_*^r > 0) \int \N(y_*^r|f_*^r,\beta_r) \, p(f_*^r|\bff^r,\bx_*^r,\{\bx_n^r\}_{n=1}^N) \, df_*^r\nonumber\\
&\approx \N(y_*^r|\mu_*^r,v_*^r).
\end{align}
This approximation is again made by moment matching, yielding
\begin{align}
\mu_*^r &= \mathbb{E}_q[f_*^r] + \sqrt{\mathbb{V}_q [f_*^r] + \beta_r} \frac{\N(z_*^r)}{\Phi(z_*^r)},\\
v_*^r &= \Big(\mathbb{V}_q[f_*^r] + \beta_r\Big) \Bigg(1 - z_*^r \frac{\N(z_*^r)}{\Phi(z_*^r)} -  \bigg(\frac{\N(z_*^r)}{\Phi(z_*^r)} \bigg)^2 \Bigg),
\end{align}
where
\begin{align}
z_*^r \triangleq \frac{\mathbb{E}_q[f_*^r]}{\sqrt{\mathbb{V}_q [f_*^r] + \beta_r}}, \nonumber
\end{align}
or using numerical quadrature for the Poisson case.
As for the equations for estimating the number of arrivals that will be caused by a given event $y_*^{e_i}$, they are analogous to the ones presented above for the number of routine arrivals. 

Finally, the predictive posterior distribution for the transportation demand (total number of arrivals) is given by
\begin{align}
p(y_*|\bx_*^r,\bX_*^e,\mathcal{D}) &= \int \N\bigg(y_* \bigg| y_*^r + \sum_{i=1}^{E_*} y_*^{e_i}, v\bigg) \, \N(y_*^r|\mu_*^r,v_*^r)\nonumber\\
&\times \prod_{i=1}^{E_*} \N(y_*^{e_i}|\mu_*^{e_i},v_*^{e_i})  \, dy_*^r \, dy_*^{e_1}\, \cdots\, dy_*^{e_{E_*}}\nonumber\\
&= \N\bigg(y_*\bigg|\mu_*^r + \sum_{i=1}^{E_*} \mu_*^{e_i}, v + v_*^r + \sum_{i=1}^{E_*} v_*^{e_i}\bigg).\nonumber
\end{align}

\section{Experiments} 
\label{section:Experiments}

The proposed Bayesian additive model with Gaussian process components (BAM-GP) was implemented in the Julia programming language\footnote{Source code is available at \url{http://fprodrigues.com/bam-src.zip}} and evaluated using both simulated data and real data from the public transport system of Singapore. 

\subsection{A toy problem}
\label{subsec:simulated}

We begin by evaluating the performance of the proposed model in a toy problem consisting of two source components A and B, characterized by feature vectors $\bx^A \sim \mbox{Unif(0,1)}$ and $\bx^B \sim \mbox{Unif(0,1)}$ respectively. Component A is assumed to be always present in all observations --- the base signal. Then, for each observation $n$, there is a variable number $E_n \sim \mbox{Poisson(1)}$ of observations from component B that are added to the base signal. 
The functions from which to obtain values of both components $y_n^A$ and $y_n^B$ are assumed to be sampled from two truncated GPs with zero mean and 0.2 signal variance. 
Both GPs use squared exponential covariance functions according to (\ref{eq:covariance_function}) with hyper-parameters $\sigma_f^2 = 2$ and $\bs\ell = \textbf{1}$. The final observations $y_n$ are then assumed to be normal distributed with mean $y_n^A + \sum_{i=1}^{E_n} y_n^{B_i}$ and 0.01 variance. 

Using this procedure, we sampled 1000 observations which were used to evaluate the proposed model with truncated Gaussian and Poisson components against the following baselines: a Bayesian linear regression that corresponds to the model in (\ref{eq:lr_simplification}), a GP where the feature vectors of component B, $\bx_n^B$, corresponding to the $n^{th}$ observation are aggregated by summation, and a version of the proposed Bayesian additive model that uses linear models with truncated Gaussian observations for the components (BAM-LR). Table~\ref{table:toy_prediction_results} shows the results obtained for predicting the observed totals $y_n$ using 10-fold cross-validation. As measures of the quality of the predictions, we report the following standard evaluation metrics: relative absolute error (RAE), correlation coefficient (CorrCoef) and coefficient of determination ($R^2$). As the results show, both versions of the proposed model outperform all the other baselines, being the the GP baseline the most competitive one.

One important advantage of considering a toy problem, is that it allows us to also evaluate the estimated values for the components $y^A$ and $y^B$ quantitatively. Table~\ref{table:toy_decomposition_results} shows the results for the posterior marginal distributions on $y^A$ and $y^B$ for the proposed model, obtained by running the EP algorithm using the entire dataset. The proposed model is compared with Bayesian linear regression, where the inferred posterior distribution of the weights $\bw$ is used to compute the marginals on the components, and with the marginal distributions obtained by running EP on BAM-LR. As the results show, the two variants of the proposed model produce the most accurate decompositions of the observed totals. Interestingly, despite the fact that the improvements of BAM-LR over a standard Bayesian linear regression in predicting the totals is only marginal, the difference in estimating the component values can be significant. This is a direct consequence of the ability of BAM-LR to properly exploit the domain knowledge in the latent additive structure of the data. 

\begin{table}[t]
\caption{Prediction results for toy problem.}
\begin{center}
\begin{tabular}{c | c c c}
Model & RAE & CorrCoef & $R^2$\\
\hline
Linear Reg. & 56.935 (1.952) & 0.795 (0.014) & 0.615 (0.028)\\
GP & 47.253 (1.511) & 0.817 (0.014) & 0.651 (0.028)\\
BAM-LR & 57.123 (1.827) & 0.802 (0.015) & 0.633 (0.026)\\
BAM-GP (poisson) & 27.371 (1.127) & 0.961 (0.003) & 0.922 (0.007)\\
BAM-GP (trunc.) & \textbf{23.468 (1.117)} & \textbf{0.971 (0.003)} & \textbf{0.941 (0.005)}
\end{tabular}
\end{center}
\label{table:toy_prediction_results}
\end{table}%

\begin{table}[t]
\caption{Decomposition results for toy problem.}
\begin{center}
\begin{tabular}{c | c | c c c}
Component & Model & RAE & CorrCoef & $R^2$ \\
\hline
\multirow{3}{*}{A} & Linear Reg. & 66.561 & 0.750 & 0.526 \\
& BAM-LR & 43.554 & 0.856 & 0.709 \\
& BAM-GP (poisson) & 21.722 & 0.967 & 0.936 \\
& BAM-GP (trunc.) & \textbf{19.864} & \textbf{0.973} & \textbf{0.946}\\
\hline
\multirow{3}{*}{B} & Linear Reg. & 98.050 & 0.339 & 0.000 \\
& BAM-LR & 69.120 & 0.667 & 0.433 \\
& BAM-GP (poisson) & 27.402 & 0.956 & 0.913 \\
& BAM-GP (trunc.) & \textbf{24.702} & \textbf{0.964} & \textbf{0.929}
\end{tabular}
\end{center}
\label{table:toy_decomposition_results}
\vspace{-0.3cm}
\end{table}%

\subsection{Real data}
\label{subsec:realdata}

The real-world data for evaluating the proposed Bayesian additive model consists of 2.7 million public transportation trip arrivals to event areas in Singapore. Singapore has a ``tap-in/tap-out'' fare card system, which provides complete origin-destination records. The dataset used contains 5 months of public transport records (from November 2012 to February 2013 and August 2013) in two popular venue areas: the Stadium and the Expo areas. Each area is served by its own subway station, whose number of arrivals we are trying to predict and dissect.

The Stadium area has two major venues: the Singapore Indoor Stadium, which is mostly home to music concerts and sports events, and the Kallang Theatre, which is a 1680-seat auditorium that usually hosts live theater performances, operas and other cultural shows. Both venues are then capable of hosting events of various types and with different target audiences. By having two co-located venues, this area allows us to understand the effect of concurrent events in close-by venues. As for the Singapore Expo, it does not have any other significant venues on the vicinity but it has a large area of $123,000$~m$^2$ with several exhibition halls. Hence, it regularly hosts multiple events at the same time (usually large exhibitions and conventions), thus making this area far more challenging to analyze.

The individual records for trips ending in one of these two areas are then aggregated by half-hour bins. Given this data, our goal is two-fold:
\begin{itemize}
\item predict the total number of arrivals by half-hour in a given area in the presence of events;
\item decompose the observed total of arrivals into the contributions of the routine component and the various events that took place in that area.
\end{itemize}
In order to achieve these goals, information about planned events is collected by mining the Web. In the next section, we will discuss how this data is collected and preprocessed. 

\subsubsection{Data preparation}

\begin{table*}[t]
\caption{Descriptive statistics for the two study areas.}
\begin{center}
\begin{tabular}{c|c|c|c|c|c|c}
Area & Total arrivals & Avg. daily arrivals & Num. events & Avg. daily events & Max. daily events & Num. days without events \\
\hline
Stadium & 570172 & 4101 ($\pm$ 925) & 34 & 0.230 ($\pm$ 0.554) & 3 & 114 (82.014\%) \\
Expo & 2088754 &15027 ($\pm$ 5515) & 342 & 2.446 ($\pm$ 1.986) & 8 & 23 (16.547\%) \\
\end{tabular}
\end{center}
\label{table:descriptivestatistics}
\end{table*}%

As previously mentioned, we consider two study areas: the Stadium and the Expo. For the five months of public transport data, a dataset of 376 events was retrieved from the Web for those 2 areas, either through screen scrapping or, when available, through the direct use of APIs. Namely, we collected events information from the following sources: \url{eventful.com}, \url{singaporeexpo.com.sg}, \url{upcoming.org}, \url{last.fm} and \url{timeoutsingapore.com}. The duplicate event titles that also share the same venue and day were merged by making use of the Jaro-Winkler string distance \cite{winkler1990string}. Table~\ref{table:descriptivestatistics} provides descriptive statistics of the collected data, which reflect some of the characteristics of the two areas and highlight their differences.

The events information consists of the title, venue, date, start time, end time, latitude, longitude, address, url, text description and categories/tags. From this information, we extract features such as the venue, whether the event has started/ended, the time to the event start/end, the event duration, if it is a multi-day event or not, etc. Since the taxonomies of the different event sources vary significantly, the categories/tags provided became hard to include in a prediction model. Alternatively, we propose the use of a web search engine in order to characterize the events according to their subject. With this aim, we use event titles and venue names as queries and then we apply a Latent Dirichlet Allocation (LDA) topic model \cite{Blei:2003:LDA:944919.944937} to the obtained search results (titles and snippets together). The number of topics in LDA was set to 25 based on an empirical analysis of the obtained topics. 
The inferred topic distributions of the different events (in form of topic weights for each event) are then used as lower-dimensional representations of their search results.

\begin{figure}[!t]
\hspace{-0.2cm}
\includegraphics[trim = 15mm 0mm 0mm 7mm, clip, scale=.53]{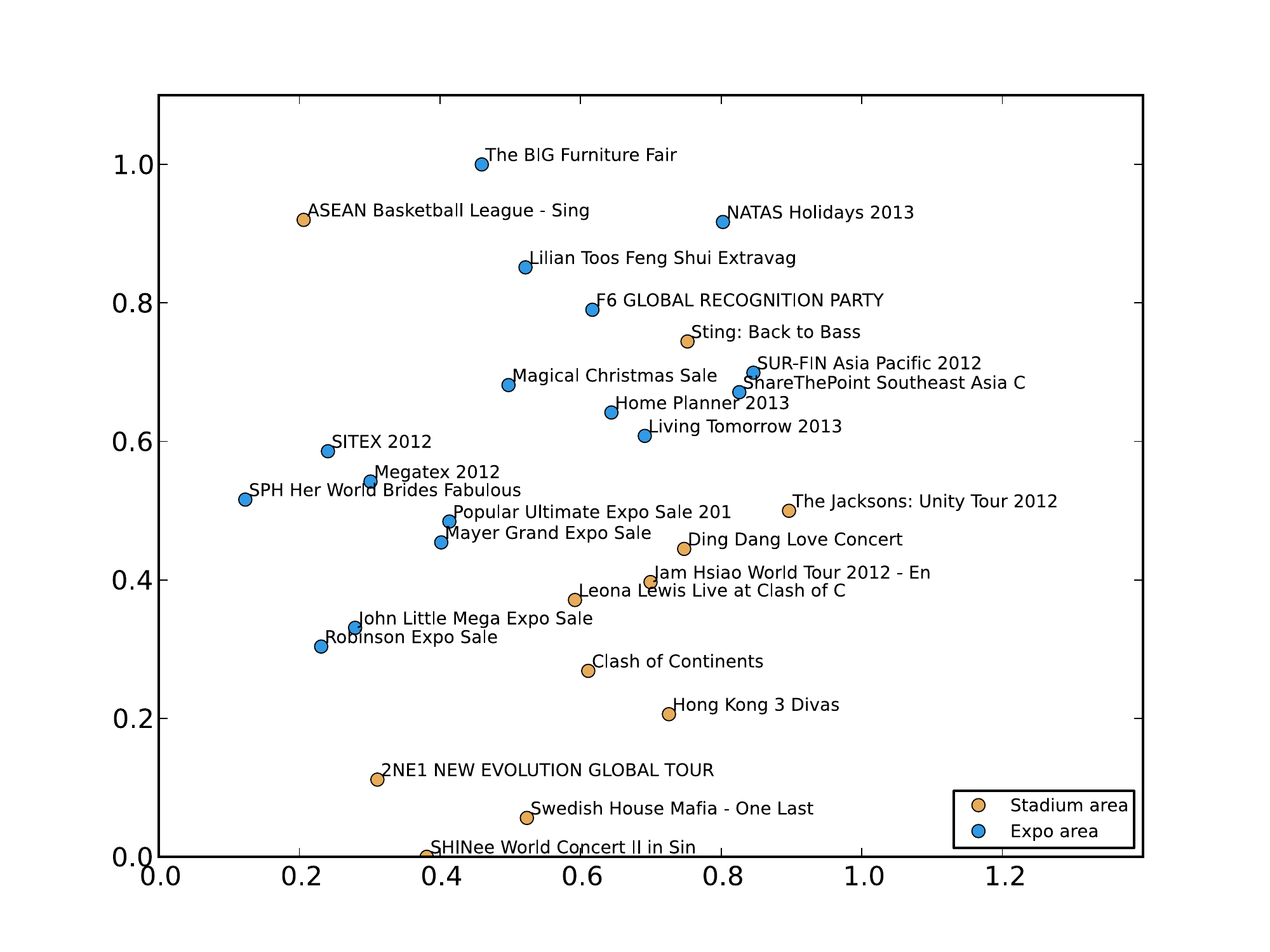}
\caption{Visualization of the topic proportions for a sample of the events data using multidimensional scaling (MDS).}
\label{fig:mds}
\end{figure}

Fig.~\ref{fig:mds} shows a 2-D visualization of the inferred topic proportions for a random sample of the events using multidimensional scaling (MDS) \cite{Borg2005}, a technique which seeks to find low-dimensional representations of the data while attempting to preserve the original distances between the data points. As the figure evidences, events with similar characteristics tend to be in the same region of the space. For example, the two electronics and IT fairs, SITEX and Megatex, are near each other. Similarly, the John Little and the Robinson (two large department stores) sales also appear together. More generally, we can notice the majority of the music-related events (\mbox{e.g.} Swedish House Mafia, SHINee, 2NE1, Leona Lewis, Jam Hsiao, etc.) being in same region of the space, separated from the rest of the events. Our hypothesis is then that events with similar topic distributions share similar effects on the observed arrivals and also on the general mobility pattern of a given place. As for the routine features, we use the weekday, time (discretized in half-hour bins) and holiday information. 

\subsubsection{Arrivals prediction}

The proposed model is evaluated using 10-fold cross validation, where the observations are ordered by time. Furthermore, the samples that belong to the same day are treated as a whole, so that they are assigned either to the test or train set altogether. This provides the model with information that is available in practice (recall that our goal with the prediction model is to make predictions far ahead of time, so that public transport operators are able to make changes accordingly). The hyper-parameters (covariance function parameters and likelihood variances) of the proposed model are determined by optimizing the marginal likelihood of the data. 
The two variants of BAM-GP, with truncated Gaussian and Poisson components, are then compared with the following baselines:
\begin{itemize}
\item two Bayesian linear regression models: one that uses only routine features, and another that corresponds to the model in Eq. (\ref{eq:lr_simplification}), which uses both routine and event features;
\item two Gaussian process models: one that considers event features and one that does not; in the case of the GP with information about events, the features of the multiple events that correspond to each observation are aggregated in the same way as with linear regression: by summing their values;
\item and a version of the proposed Bayesian additive model that uses linear models for the routine and the events components (BAM-LR). This approach is generally similar to the one used in \cite{pereira2015so}. 
\end{itemize}
The hyper-parameters of all the baselines (prior and likelihood hyper-parameters of the linear models and GP's covariance function hyper-parameters) were also set by optimizing the marginal likelihood of their respective models.

\begin{table*}[!t]
\caption{Results (with standard errors) for estimating the total arrivals in the Stadium area using 10-fold cross-validation.}
\begin{center}
\begin{tabular}{c | c | c c c | c c c}
&& \multicolumn{3}{c|}{Evaluation: all times} & \multicolumn{3}{c}{Evaluation: event periods only}\\
\hline
Area & Model & RAE & CorrCoef & $R^2$ & RAE & CorrCoef & $R^2$ \\
\hline
\multirow{7}{*}{\rotatebox[origin=c]{0}{Stadium}}&Linear Reg. (routine only) & 52.652 (1.587) & 0.752 (0.022) & 0.542 (0.032) & 66.817 (8.322) & 0.783 (0.065) & 0.457 (0.108) \\
&Linear Reg. (routine + events) & 48.368 (1.049) & 0.822 (0.017) & 0.655 (0.034) & 55.493 (7.710) & 0.864 (0.025) & 0.627 (0.057) \\
&GP (routine only) & 49.191 (2.231) & 0.777 (0.026) & 0.580 (0.039) & 62.069 (8.584) & 0.801 (0.066) & 0.409 (0.105) \\
&GP (routine + events) & 46.357 (2.393) & 0.833 (0.033) & 0.675 (0.059) & 59.761 (9.876) & 0.802 (0.080) & 0.625 (0.056) \\
&BAM-LR & 48.315 (1.225) & 0.815 (0.020) & 0.641 (0.038) & 54.242 (7.601) & 0.859 (0.029) & 0.613 (0.080) \\
&BAM-GP (poisson) & 44.240 (2.275) & 0.859 (0.029) & 0.718 (0.058) & 45.504 (9.893) & 0.894 (0.031) & 0.788 (0.029) \\
&BAM-GP (truncated) & \textbf{42.222 (1.819)} & \textbf{0.871 (0.021)} & \textbf{0.745 (0.039)} & \textbf{42.338 (6.172)} & \textbf{0.907 (0.022)} & \textbf{0.789 (0.034)} \\
\hline
\multirow{7}{*}{\rotatebox[origin=c]{0}{Expo}}&Linear Reg. (routine only) & 63.487 (2.603) & 0.745 (0.019) & 0.517 (0.040) & 82.999 (6.354) & 0.537 (0.056) & 0.323 (0.051) \\
&Linear Reg. (routine + events) & 59.480 (3.863) & 0.791 (0.020) & 0.544 (0.059) & 81.029 (5.749) & 0.620 (0.052) & 0.370 (0.048) \\
&GP (routine only) & 40.096 (3.163) & 0.898 (0.016) & 0.765 (0.045) & 56.473 (6.638) & 0.843 (0.038) & 0.698 (0.046) \\
&GP (routine + events) & 37.128 (1.702) & 0.893 (0.013) & 0.771 (0.027) & 54.743 (4.034) & 0.798 (0.037) & 0.576 (0.074) \\
&BAM-LR & 61.521 (3.213) & 0.762 (0.028) & 0.542 (0.050) & 77.928 (5.697) & 0.589 (0.057) & 0.355 (0.059) \\
&BAM-GP (poisson) & 46.822 (2.667) & 0.891 (0.014) & 0.745 (0.034) & 50.783 (3.868) & 0.873 (0.026) & 0.684 (0.058) \\
&BAM-GP (truncated) & \textbf{33.411 (2.040)} & \textbf{0.927 (0.013)} & \textbf{0.835 (0.034)} & \textbf{46.033 (4.569)} & \textbf{0.884 (0.032)} & \textbf{0.720 (0.077)}
\end{tabular}
\end{center}
\label{table:results_crossval} 
\end{table*}

\begin{figure*}[!t]
\centering
\subfloat[]{\includegraphics[width=0.4\linewidth,trim = 10mm 50mm 30mm 5mm]{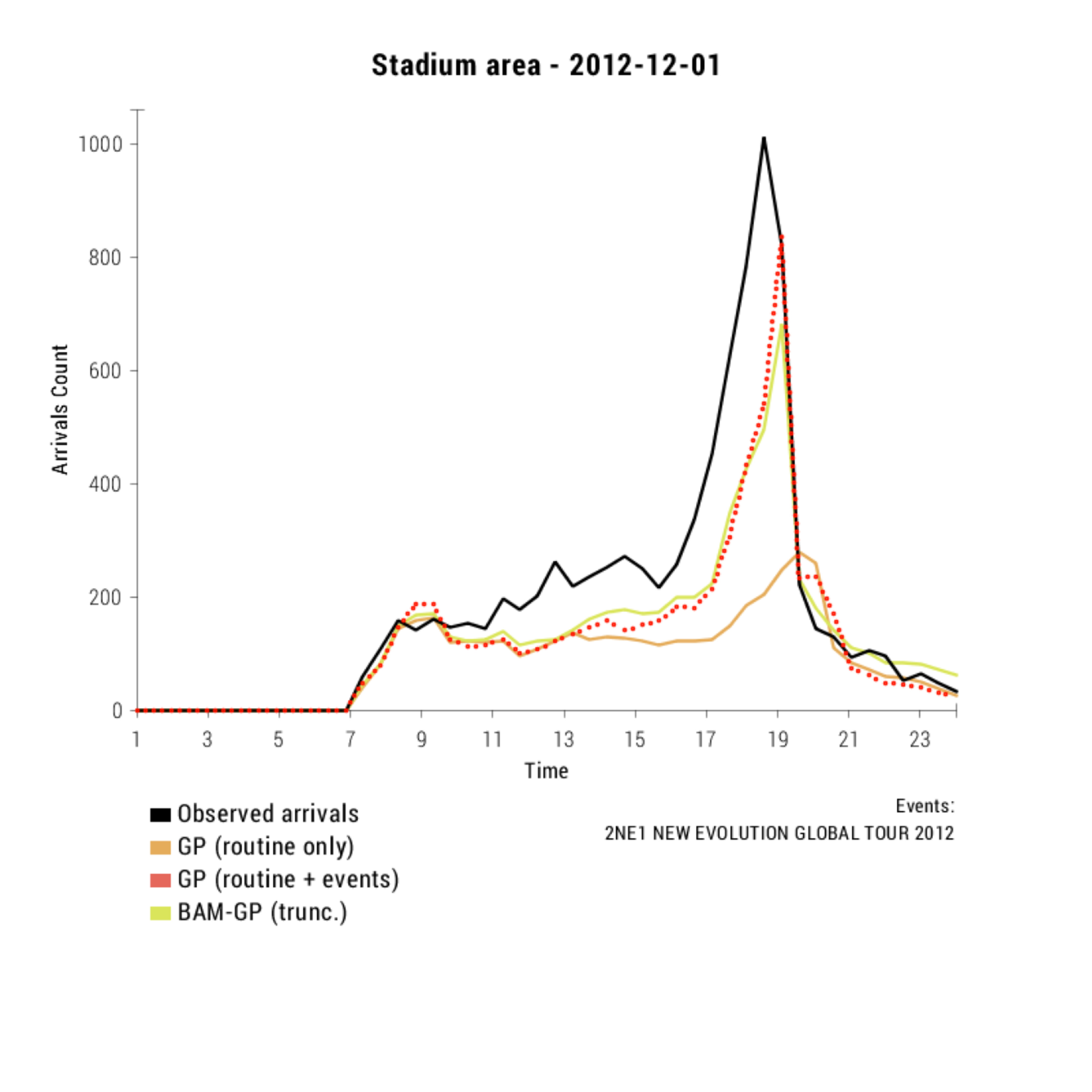}\label{fig:predictions1}}\hspace*{0.5cm}
\subfloat[]{\includegraphics[width=0.4\linewidth,trim = 10mm 50mm 30mm 5mm]{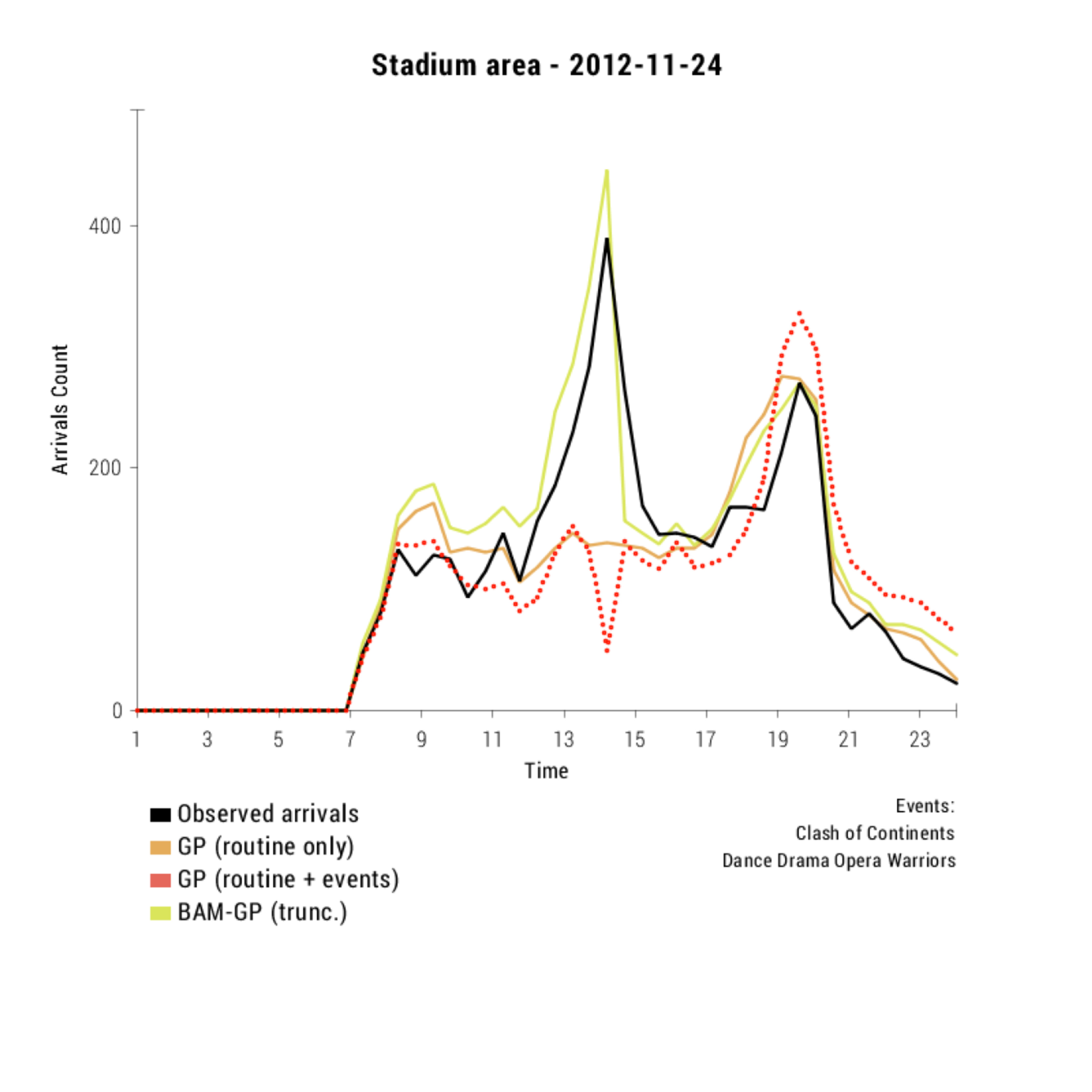}\label{fig:predictions2}}
\vspace{0.3cm}
\subfloat[]{\includegraphics[width=0.4\linewidth,trim = 10mm 20mm 30mm 5mm]{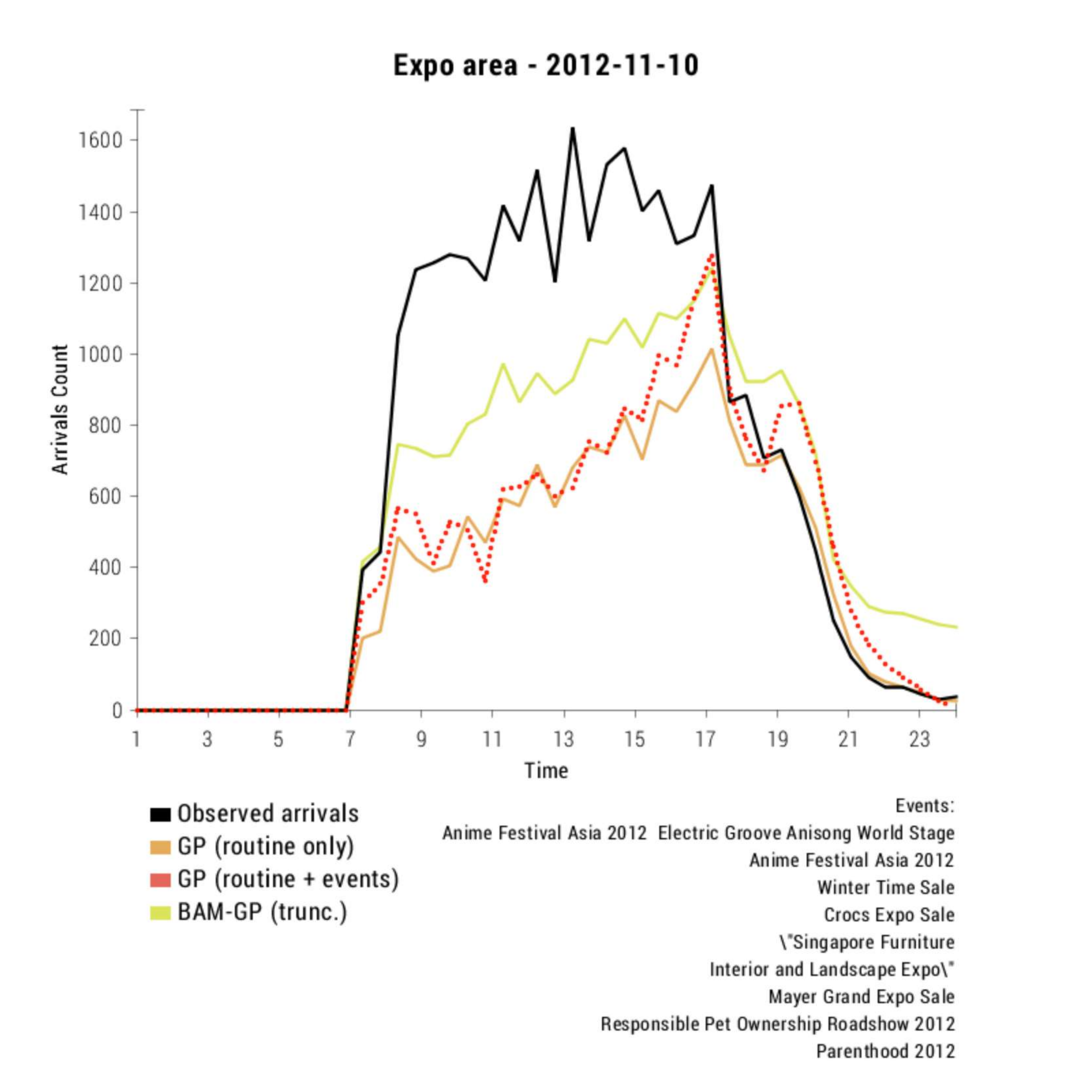}\label{fig:predictions3}}\hspace*{0.5cm}
\subfloat[]{\includegraphics[width=0.4\linewidth,trim = 10mm 20mm 30mm 5mm]{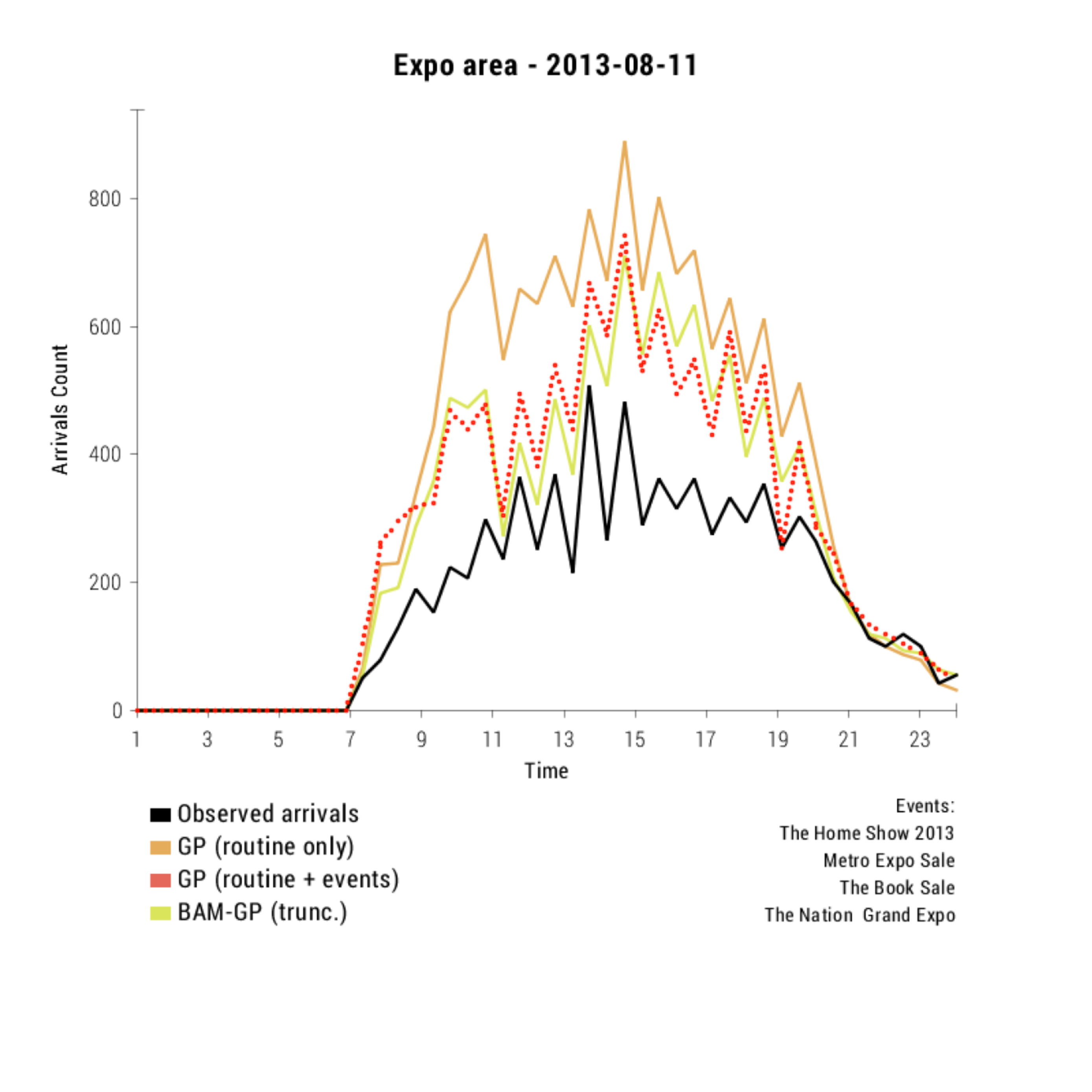}\label{fig:predictions4}} 
\caption{Comparison of the predictions of BAM-GP (with truncated Gaussians) with the true observed arrivals (black solid line) and the predictions of the GP models for four example days.}
\label{fig:prediction_set1}
\end{figure*}

\begin{figure*}[!t]
\centering
\subfloat[Linear Reg. (routine + events)]{\includegraphics[width=0.36\linewidth,trim = 13mm 45mm 10mm 5mm,clip]{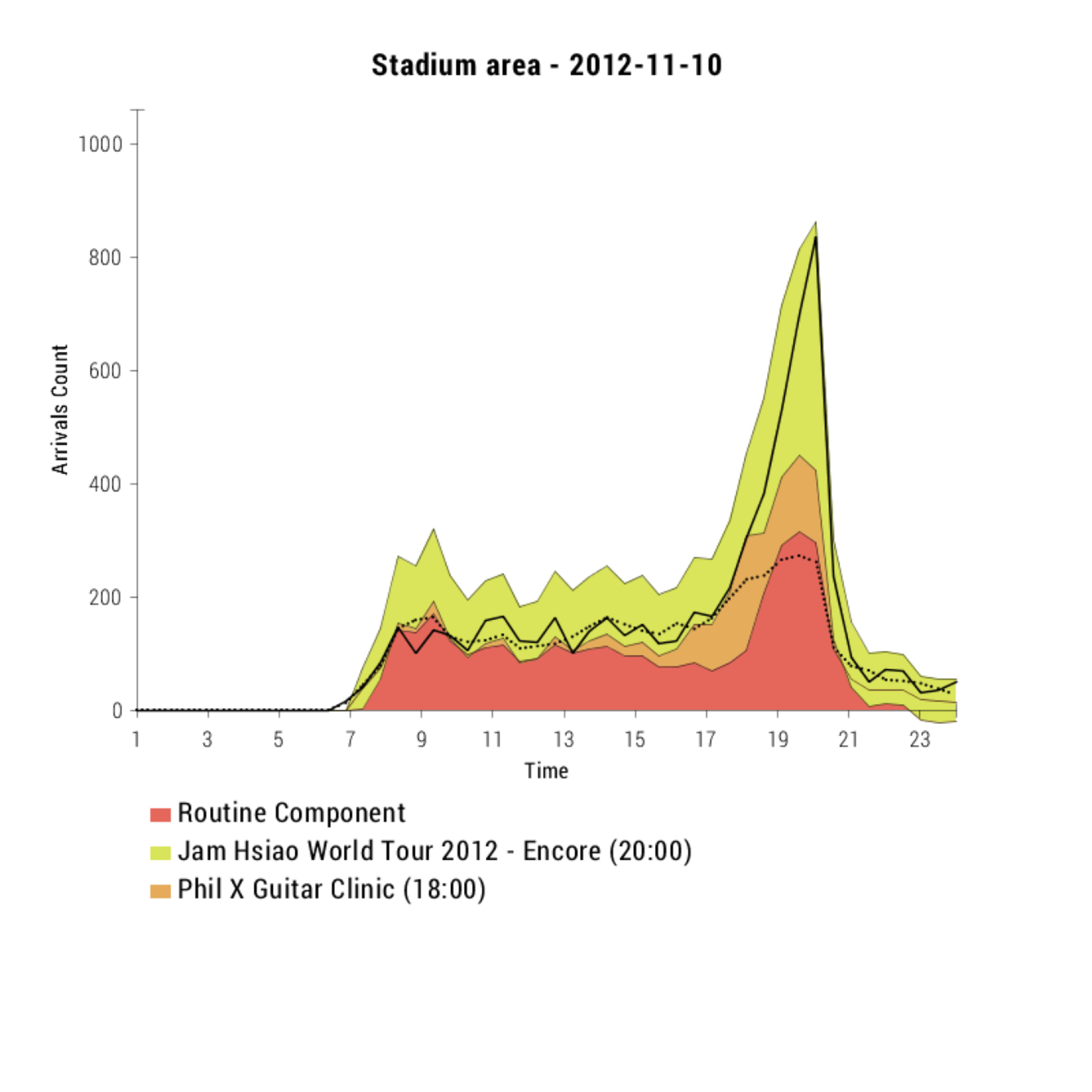}\label{fig:marginals_demo1_lr}}\hspace*{-0.5cm}
\subfloat[BAM-LR]{\includegraphics[width=0.36\linewidth,trim = 13mm 45mm 10mm 5mm,clip]{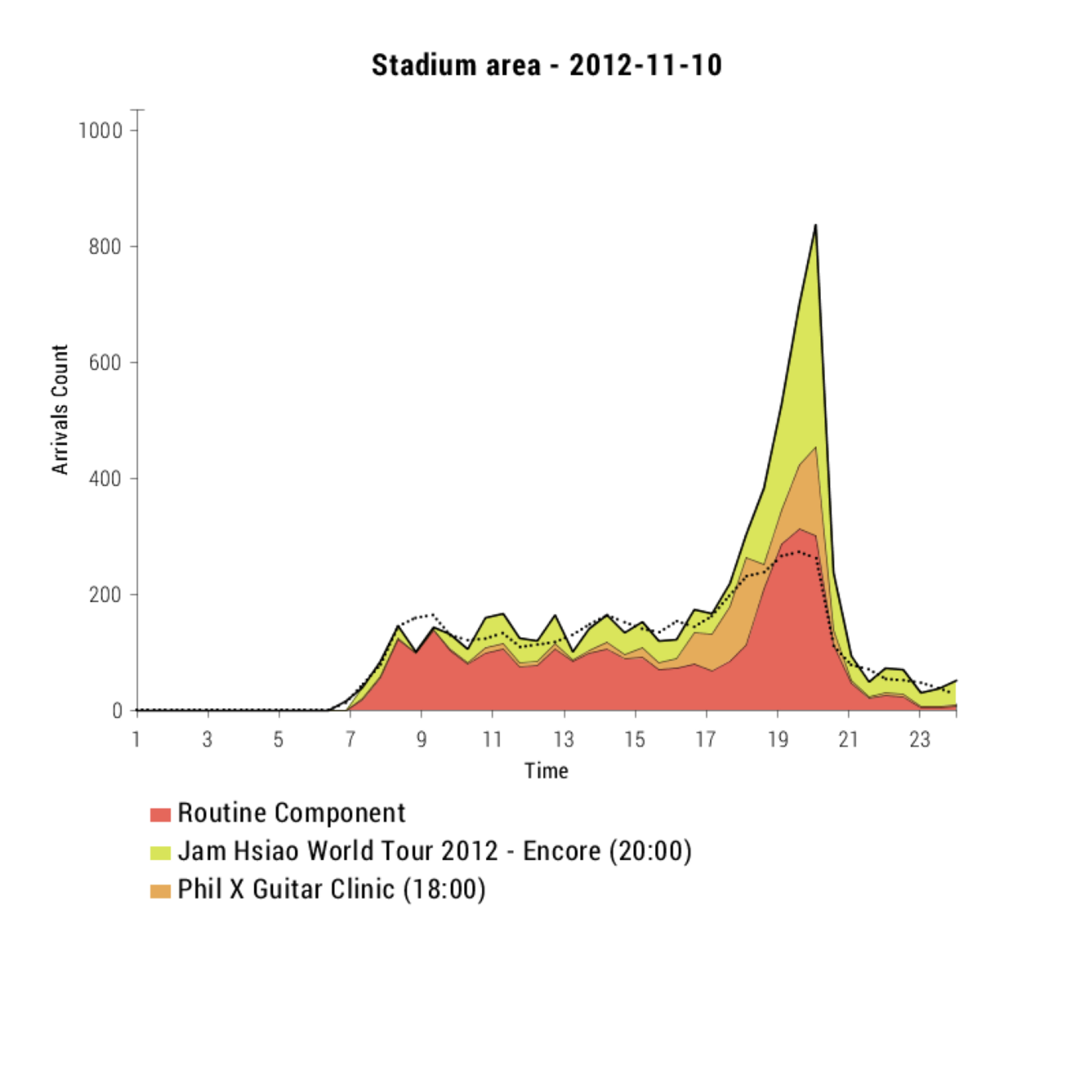}\label{fig:marginals_demo1_balm}}\hspace*{-0.5cm}
\subfloat[BAM-GP]{\includegraphics[width=0.36\linewidth,trim = 13mm 45mm 10mm 5mm,clip]{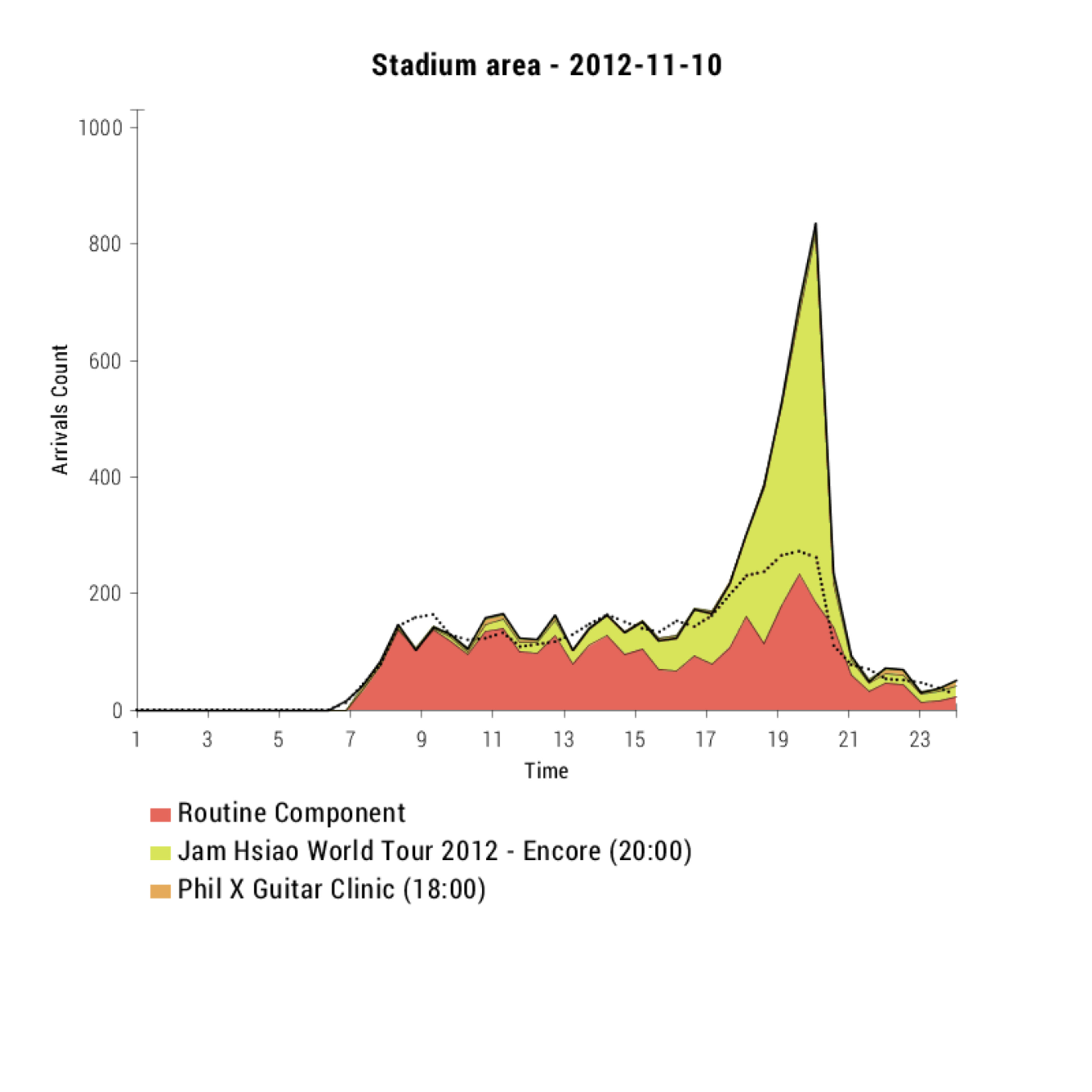}\label{fig:marginals_demo1_bam}}

\vspace{0.4cm}
\subfloat[Linear Reg. (routine + events)]{\includegraphics[width=0.36\linewidth,trim = 13mm 15mm 10mm 5mm,clip]{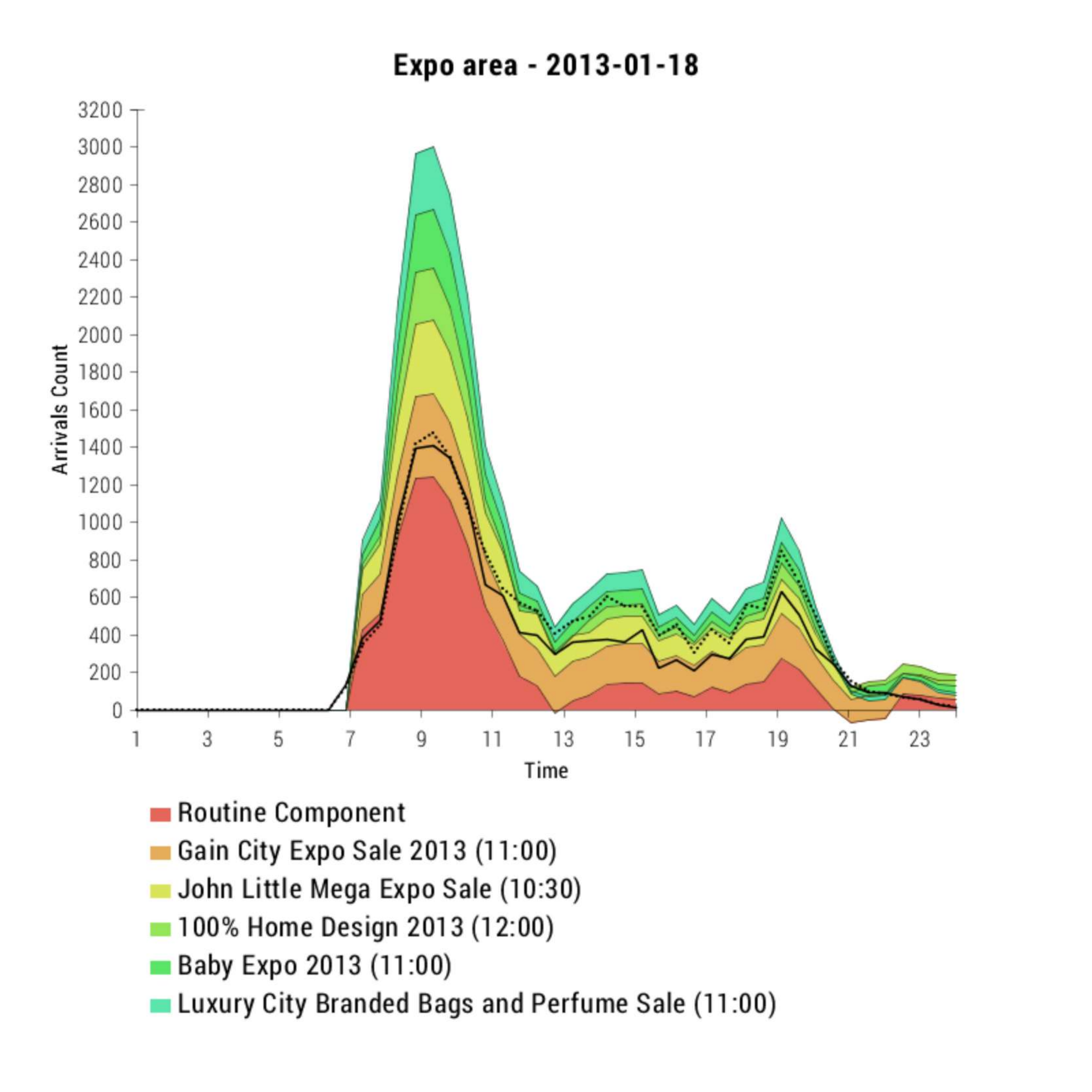}\label{fig:marginals_demo2_lr}}\hspace*{-0.5cm}
\subfloat[BAM-LR]{\includegraphics[width=0.36\linewidth,trim = 13mm 15mm 10mm 5mm,clip]{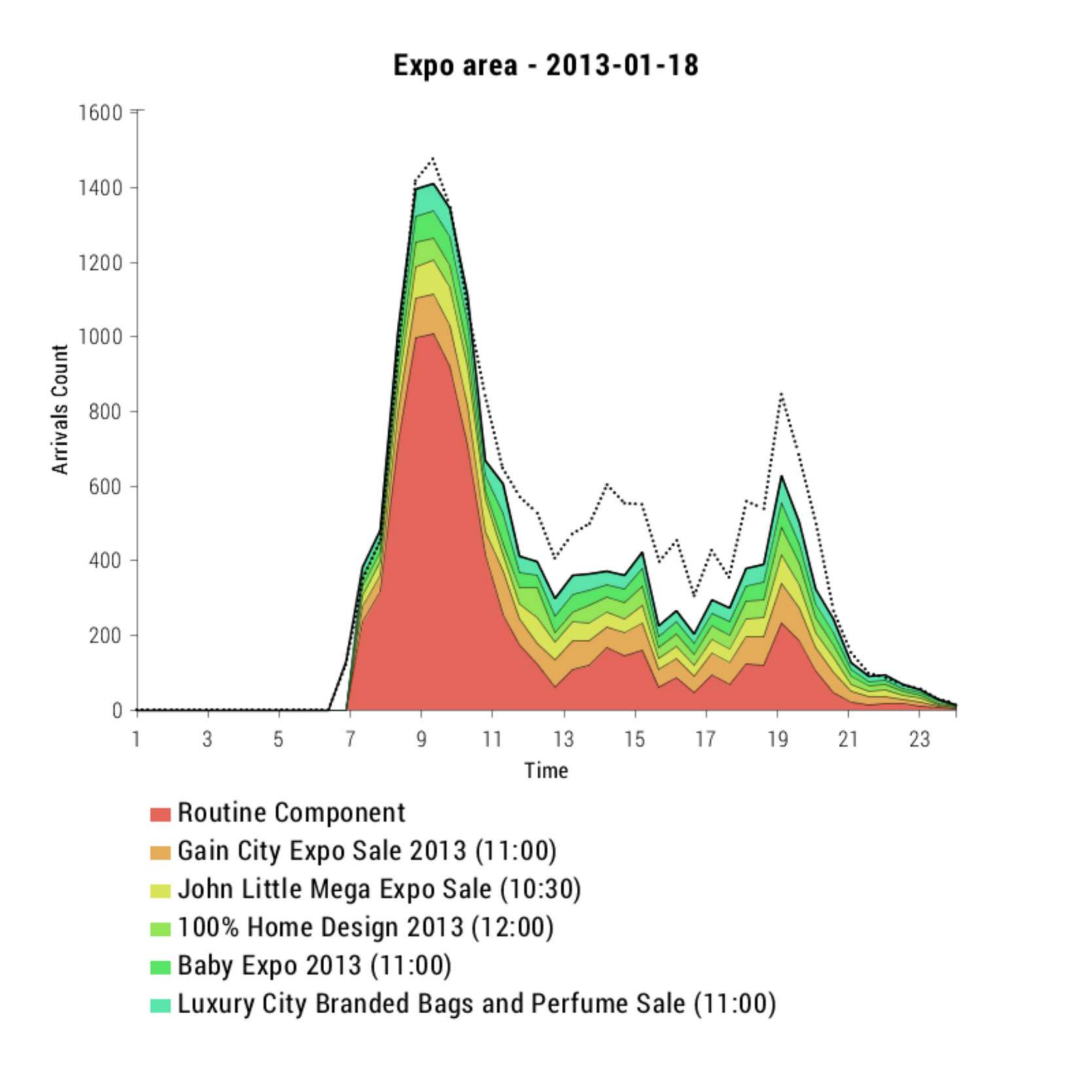}\label{fig:marginals_demo2_balm}} \hspace*{-0.5cm}
\subfloat[BAM-GP]{\includegraphics[width=0.36\linewidth,trim = 13mm 15mm 10mm 5mm,clip]{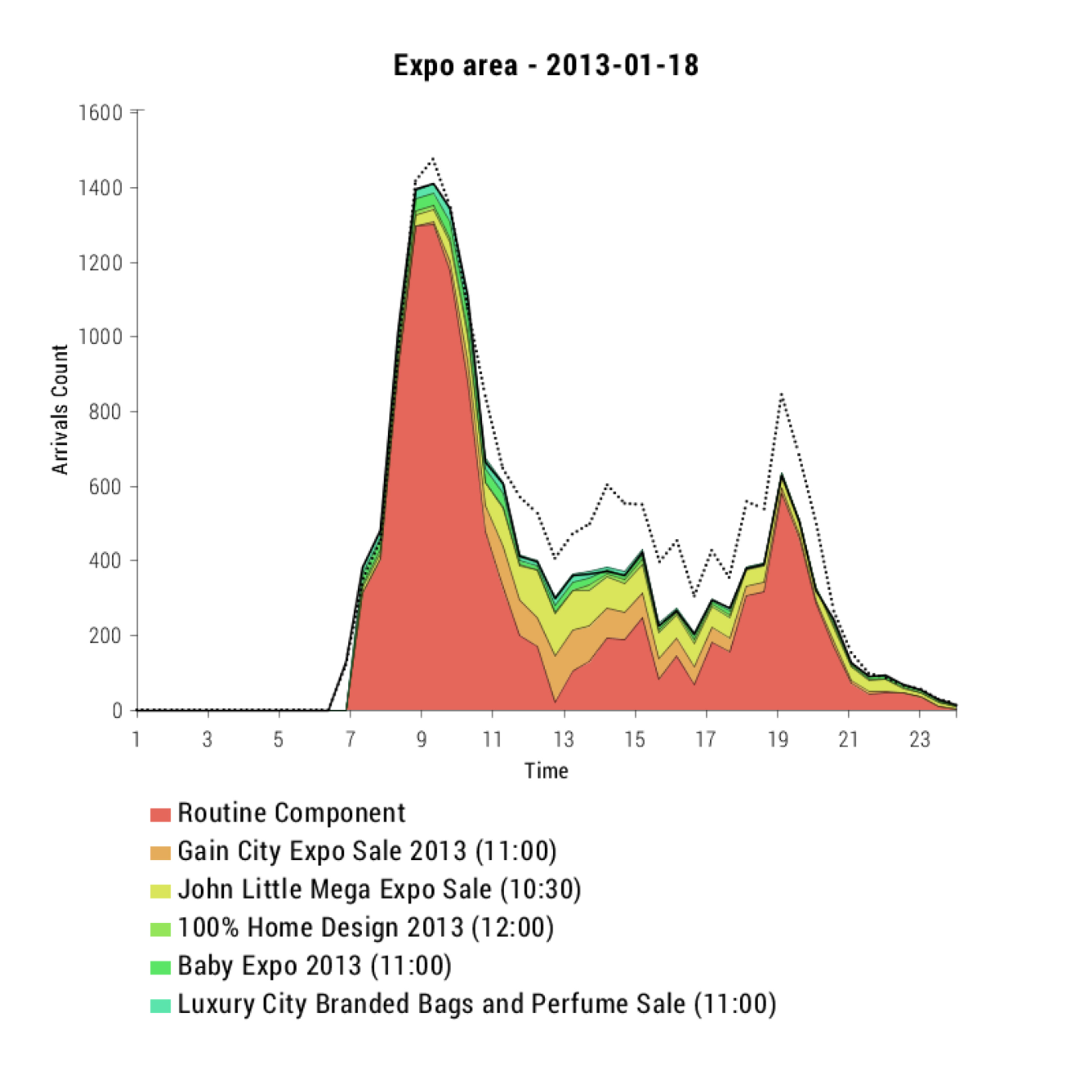}\label{fig:marginals_demo2_bam}} 
\caption{Results obtained by 3 different approaches (columns) on two examples days in two different areas (rows) for disaggregating the total observed arrivals (black solid line) into the contributions of the routine component and the various nearby events. The dotted line represents the median arrivals over all the days in the observed data that correspond the same weekday. Events start times are shown in parentheses.}
\label{fig:decomposition_comparison}
\end{figure*} 

Table~\ref{table:results_crossval} shows the results obtained. Besides a global evaluation, we also provide error metrics only for the periods when events are about to start (time to event start is less than an hour) or ongoing only, so that the contribution of the models that include information regarding events can be more evident. 

Let us start by analysing the value of including events information in standard approaches such as Bayesian linear regression and Gaussian processes. The obtained results clearly show the advantage of incorporating such information, being the improvements particularly noticeable in the Stadium area. This is not surprising since the Stadium area has an average of 0.230 events per day, which makes it easier to add additional features to the input vectors of these models in order to account for the effect of events. However, in Expo, the number of events can go up to 8 in a single day (see Table~\ref{table:descriptivestatistics}). This makes the feature aggregation problem described in Section~\ref{section:problem_fromulation} more severe. This is especially visible in the results of the GPs for Expo, where we can see that including events information leads only to marginal overall improvements over the GP that just considers routine information. Furthermore, when we focus the evaluation on event periods only, we can see that including this information actually leads to worse results. 

Regarding the comparison between linear and non-linear approaches, the results from Table~\ref{table:results_crossval} show that the GPs outperform their linear regression counterparts, thus motivating the need for non-linear models. This difference is particularly clear in Expo, where using a GP instead of linear regression can lead to error reductions up to 38\% in RAE. 

Finally, the proposed BAM-GP with truncated Gaussian components is shown to outperform all the other methods in both areas, being the GP model that uses routine and event features the closest baseline method. 
This is further evidence that fusing data at model-level leads to better results than doing so at feature-level \cite{zheng2014urban}. 
Interestingly, the difference between BAM-GP and BAM-LR is shown to be very significant, which once more justifies the use of non-linear models and the assumption of GP priors over the latent function values, $\bff^r$ and $\bff^e$, of the components. 

As for the comparison between the use of truncated Gaussian or Poisson distributions to model the contributions of the component $y_n^r$ and $y_n^{e_i}$, the results in Table~\ref{table:results_crossval} clearly show that the use of truncated Gaussians leads to better results (the statistical difference was verified at the 0.05 level). This conforms with preliminary distribution fitting tests that we performed. Furthermore, the use of truncated Gaussians has the advantage that the moments' calculations required for EP can be done analytically, while the Poisson case requires the use of more expensive quadrature procedures. 

\begin{table}[t]
\caption{Top-10 more relevant features according to ARD.}
\begin{center}
\begin{tabular}{c | c | l}
Area&$\ell$&Feature\\
\hline
\multirow{10}{*}{\rotatebox[origin=c]{90}{Stadium}}&2.490&Time to event start\\
&5.398&Topic 22: world, tour, concert, girl, gener\\
&6.651&Topic 21: super, tour, world, junior, show\\
&8.262&Topic 14: cirqu, soleil, saltimbanco, iron, maiden\\
&9.142&Venue = Singapore Indoor Stadium?\\
&9.681&Is Friday?\\
&10.106&Topic 4: music, rock, fan, song, kim\\
&12.465&Topic 19: basketbal, asean, leagu, slinger, abl\\
&13.286&Day without events?\\
&14.297&Topic 18: direct, nile, mariah, rodger, carey\\
\hline
\multirow{10}{*}{\rotatebox[origin=c]{90}{Expo}}&2.540&Topic 24: sale, shop, great, warehous, deal\\
&4.026&Topic 13: electron, citi, gain, consum, show\\
&5.092&Time to event start\\
&7.638&How long has the event started\\
&11.142&Topic 9: asia, pacif, confer, exhibit, industri\\
&11.346&Topic 25: food, travel, fair, natas, halal\\
&11.349&Topic 11: properti, real, invest, trade, estate\\
&13.980&Event has started?\\
&14.058&Topic 4: beauti, facebook, john, product, loreal\\
&14.580&Topic 7: home, busi, job, global, nation\\
\end{tabular}
\end{center}
\label{table:top10_features}
\vspace{-0.4cm}
\end{table}%

From the perspective of public transport operators and planners, it is important to note that, in practice, the typical approach is to rely on historical averages or simple linear models that do not account for event information. Therefore, the proposed Bayesian additive model with GP components constitutes a significant contribution in comparison to current practice. Even in cases where these rely on more complex models such as GPs, the use of BAM-GP and the inclusion of event information can lead to improvements in $R^2$ by as much as 92\%, as shown for the Stadium area during event periods. Improvements of this magnitude are expectable to produce very significant effects in practice. 

In order to illustrate some of these differences in their real-world context, we plotted in Fig.~\ref{fig:prediction_set1} the predictions of BAM-GP with truncated Gaussians and the GP models against the true observed arrivals for four example days. The obtained results further highlight the practical implications of the improvements obtained by the proposed additive model. For example, in Fig.~\ref{fig:predictions1} the GP model that only uses routine information underestimates the huge peek in arrivals due to the big concert by 2NE1 --- a very popular band in the southeast Asia. As the figure shows, the models that also incorporate event information produce much better predictions. Figs.~\ref{fig:predictions2} and \ref{fig:predictions3} show two additional examples where the GP that only considers routine information fails to predict the increased demand caused by the events. However, possibly due to the multiple concurrent events, the inclusion of event information in the GP does not seem to improve the predictions. In both cases, the proposed BAM-GP provides much more accurate predictions. Lastly, Fig.~\ref{fig:predictions4} shows a case where the routine GP actually overestimates the demand, while the models that include event information perform much better.

With the purpose of understanding what the additive models are doing internally, Table~\ref{table:top10_features} shows the top-10 features for which the length-scales of the ARD covariance function are smaller, for the Stadium and Expo areas respectively. Interestingly, the tables show that the more relevant features for the Stadium area are the topics that characterize big concerts and the time for their beginning, while the most relevant features for the Expo are topics related to big sales and electronic exhibits, which conforms with the well-known shopping and tech-enthusiastic culture of Singapore.

\subsubsection{Arrivals decomposition}

\begin{figure*}[h!]
\centering
\subfloat[]{\includegraphics[width=0.36\linewidth,trim = 13mm 45mm 10mm 5mm,clip]{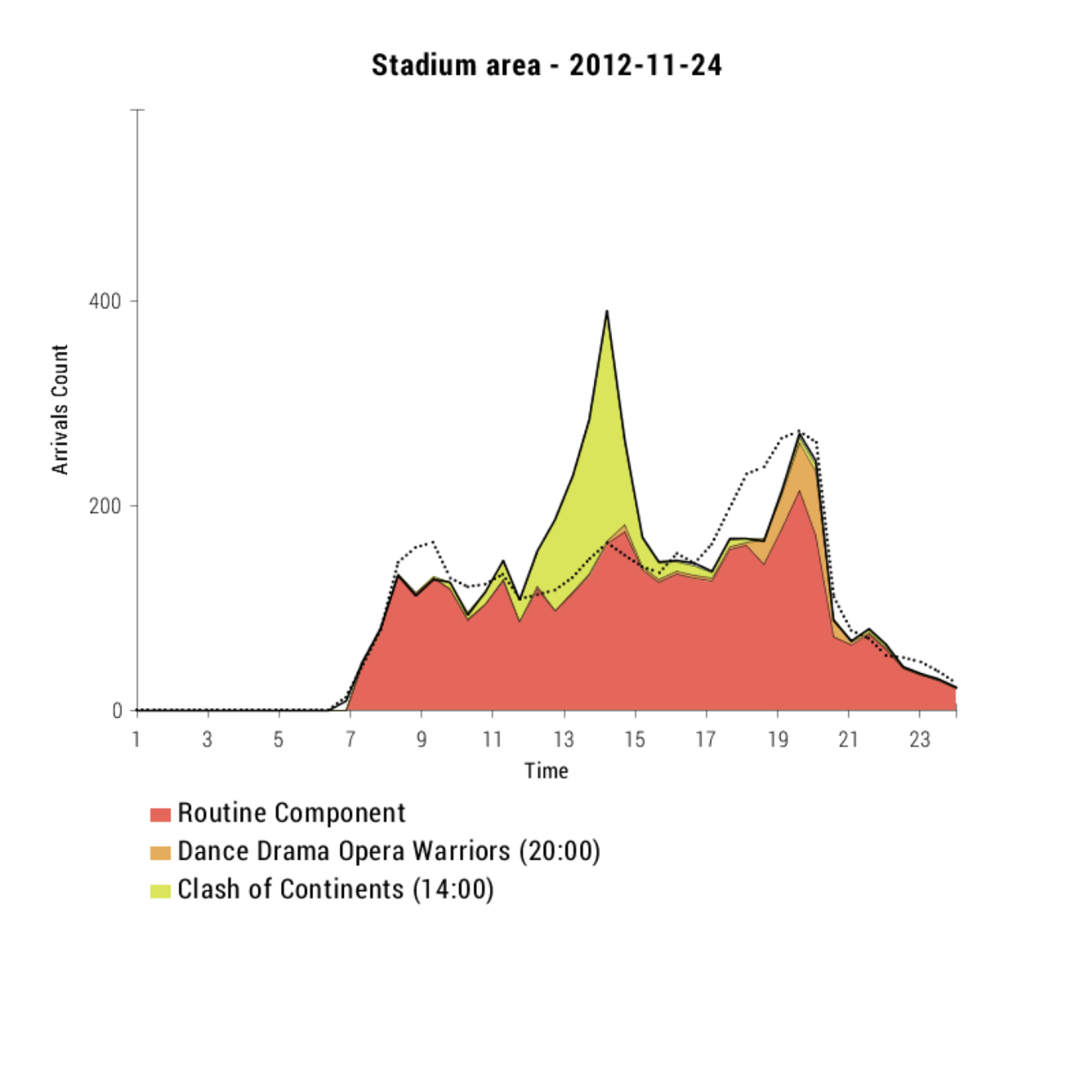}\label{fig:marginals1}}\hspace*{-0.5cm}
\subfloat[]{\includegraphics[width=0.36\linewidth,trim = 13mm 45mm 10mm 5mm,clip]{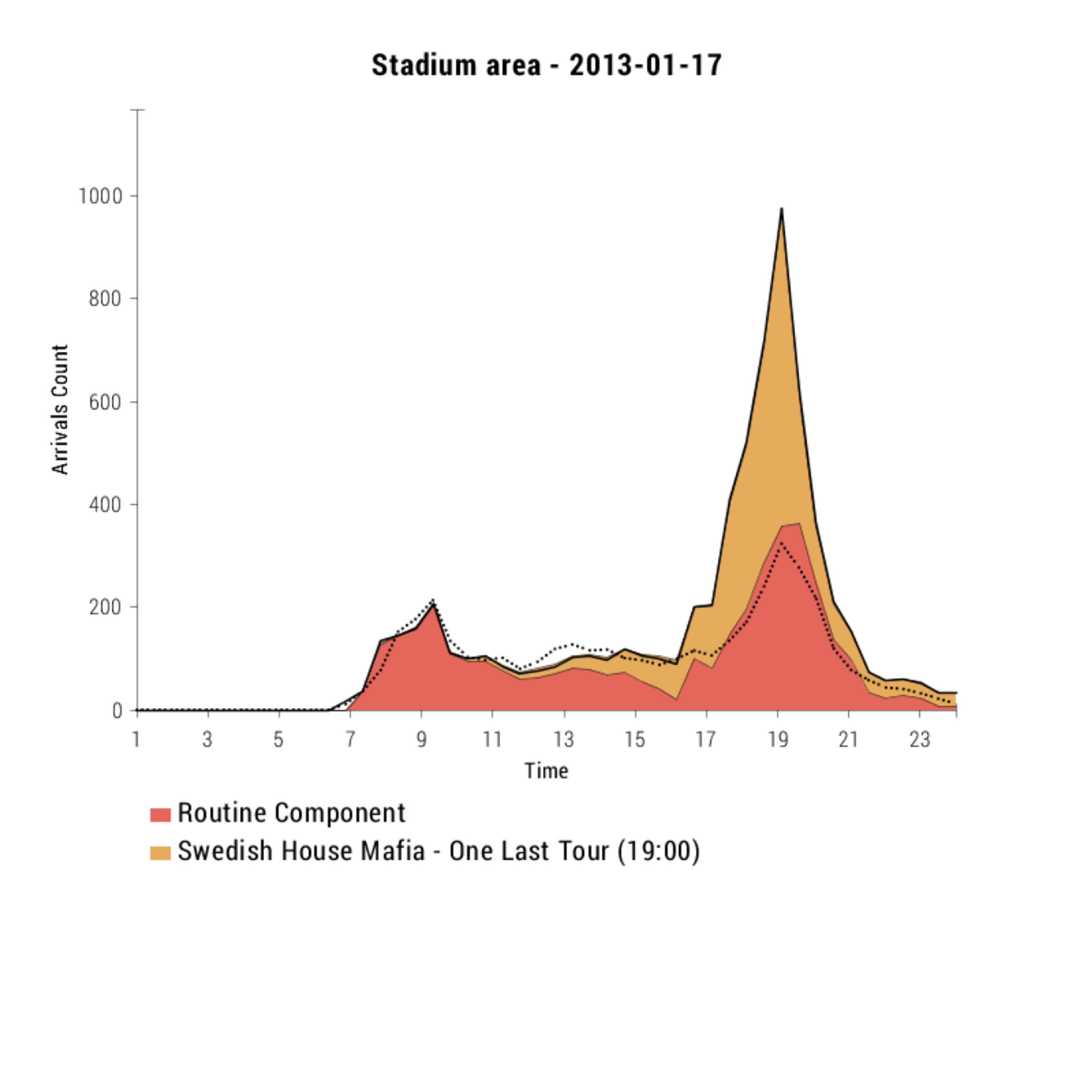}\label{fig:marginals2}}\hspace*{-0.5cm}
\subfloat[]{\includegraphics[width=0.36\linewidth,trim = 13mm 45mm 10mm 5mm,clip]{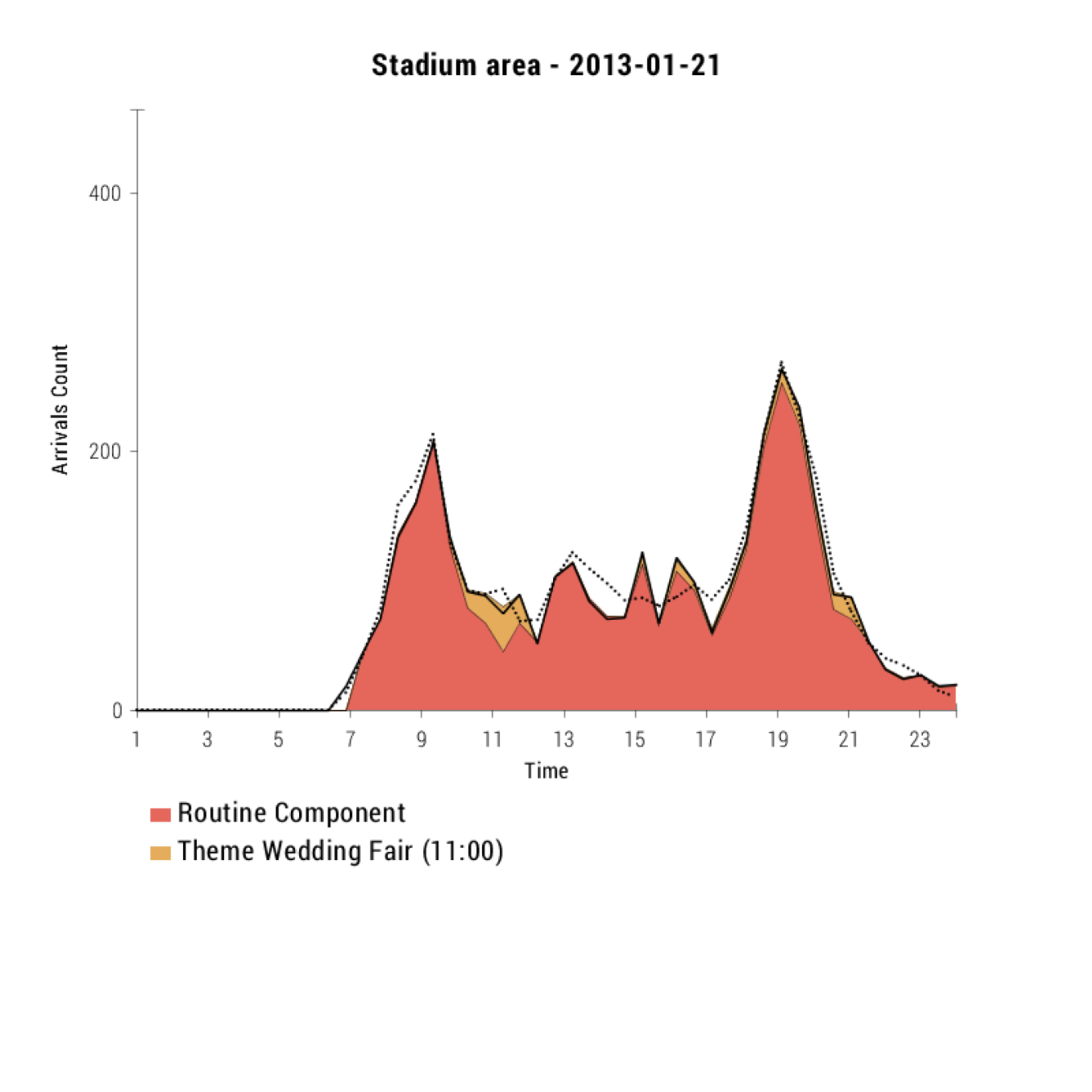}\label{fig:marginals3}}

\vspace{0.4cm}
\subfloat[]{\includegraphics[width=0.36\linewidth,trim = 13mm 0mm 10mm 5mm,clip]{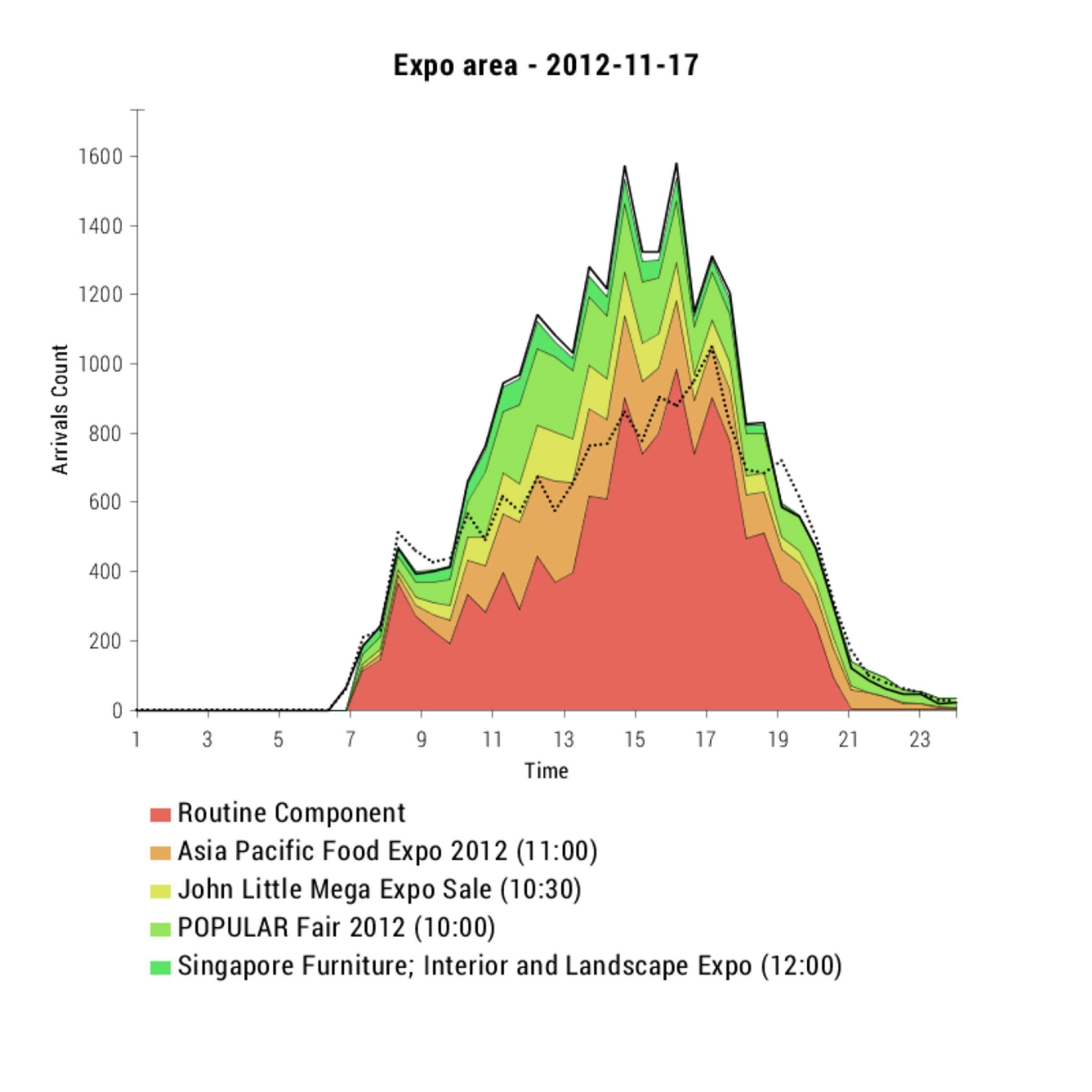}\label{fig:marginals4}}\hspace*{-0.5cm}
\subfloat[]{\includegraphics[width=0.36\linewidth,trim = 13mm 0mm 10mm 5mm,clip]{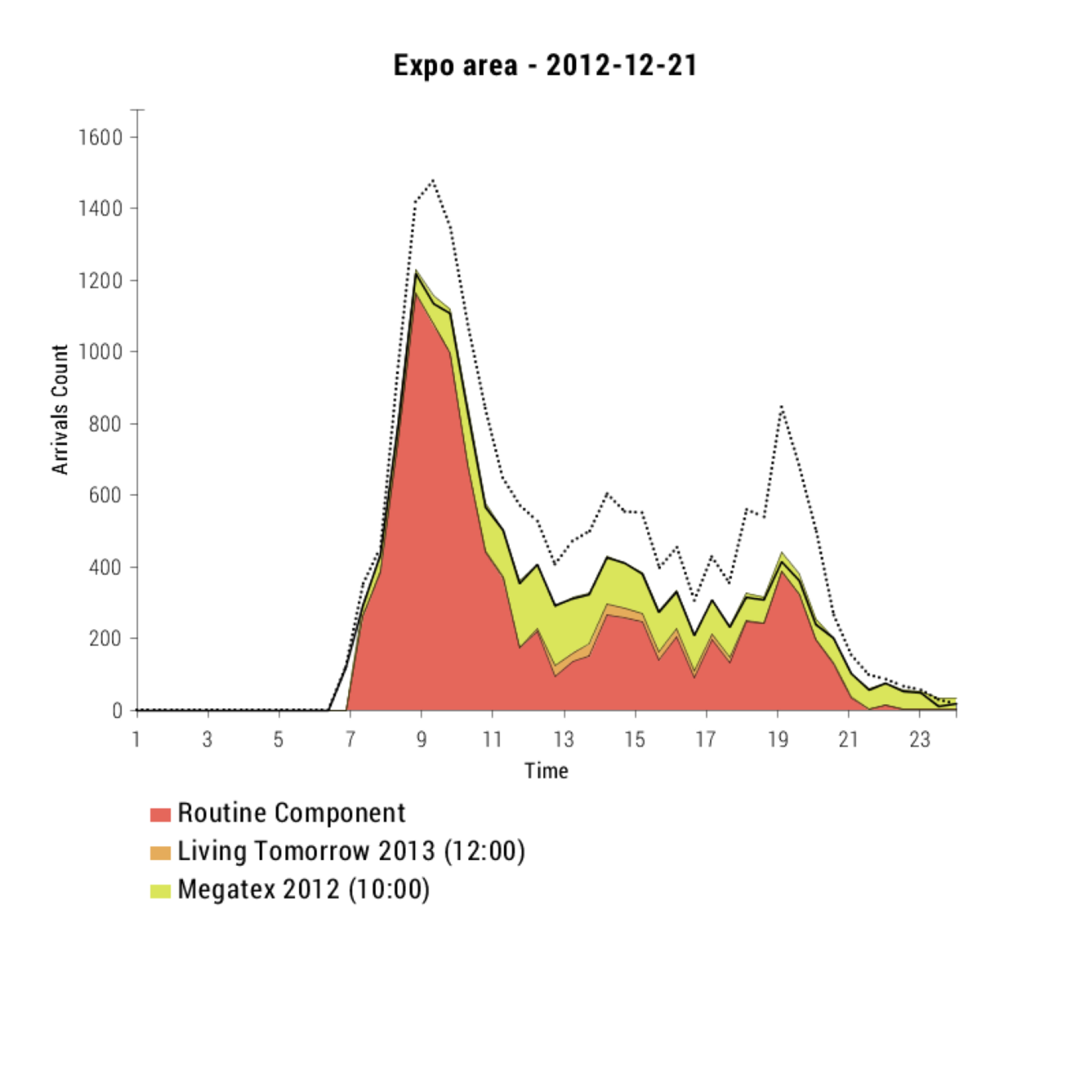}\label{fig:marginals5}} \hspace*{-0.5cm}
\subfloat[]{\includegraphics[width=0.36\linewidth,trim = 13mm 0mm 10mm 5mm,clip]{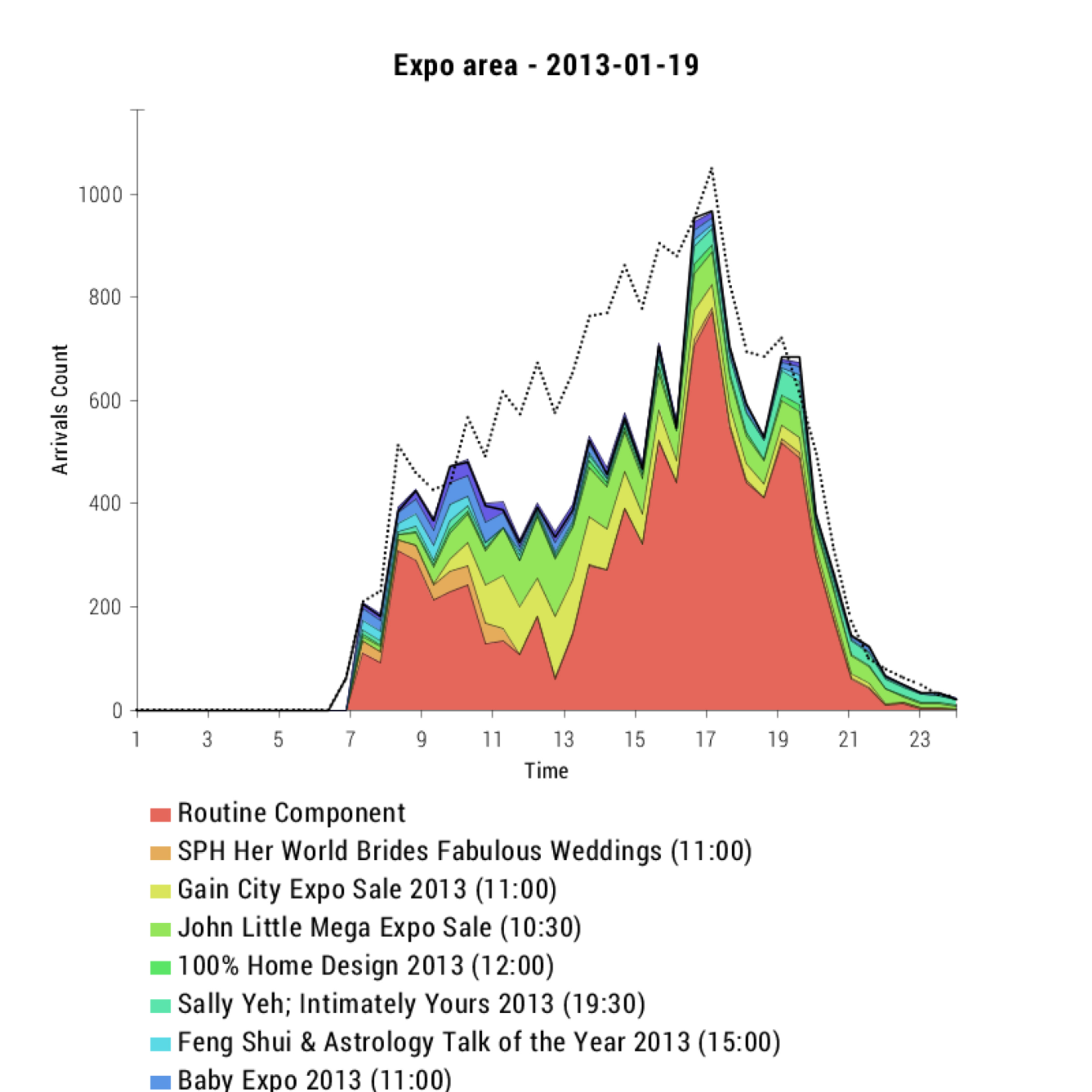}\label{fig:marginals6}} 
\caption{Results obtained by BAM-GP for disaggregating the total observed arrivals (black solid line) in 6 example days into the contributions of the routine component and the various nearby events. The dotted line represents the median arrivals over all the days in the observed data that correspond to the same weekday. Events start times are shown in parentheses.}
\label{fig:decomposition_bam}
\end{figure*}

In order to analyze decomposition results generated by the additive model, we need to take a closer look at the posterior marginal distributions on the routine component $y^r$ and on the events components $\{y^{e_i}\}_{i=1}^E$ estimated by BAM-GP by running the EP inference algorithm on the entire dataset. Since it is impossible to obtain ground truth for this particular decomposition problem, our analysis will be qualitative rather than quantitative, and more based on common sense. 
With this aim, performance of BAM-GP with truncated Gaussian components will be compared with the linear regression model in (\ref{eq:lr_simplification}), where the inferred posterior distribution of the weights $\bw$ is used to compute the posterior on the components $y^r$ and $\{y^{e_i}\}_{i=1}^E$ for each individual observation by making use of Bayes' theorem, and with BAM-LR.

Fig.~\ref{fig:decomposition_comparison} shows the results obtained by the three different approaches (columns) for two illustrative example days in the two study areas (rows). Let us start by analyzing the first row of examples, which correspond to the 10th of November 2012 in the Stadium area. From Fig.~\ref{fig:marginals_demo1_lr} it can be seen that the component values estimated by the linear regression model do not add up the total observed arrivals, which makes the output of this approach harder to use in practice. The BAM-LR decomposition from Fig.~\ref{fig:marginals_demo1_balm} matches closer to the observed totals, however it estimates a significant amount of arrivals caused by the Jam Hsiao concert long after that event is over, which is unlikely. Fig.~\ref{fig:marginals_demo1_bam} shows that BAM-GP not only overcomes those problems, but it also provides more intuitive results. It assigns almost all the demand to the big concert by Jam Hsiao, a young star mandopop singer widely popular in Asian counties, that took place in the Indoor Stadium, as opposed to the small guitar clinic (a masterclass workshop for rock guitar players). 

The results for the Expo area (Fig.~\ref{fig:marginals_demo2_lr}) illustrate another weakness of the linear regression model: by not incorporating any constraints on the components, the estimated number of arrivals around 21:00 due to routine commuting becomes negative. By assuming a truncated Gaussian distribution for the components distributions, the Bayesian additive models do not suffer from this problem. However, as Fig.~\ref{fig:marginals_demo2_balm} evidences, the simpler BAM-LR model is once again suffering from the problem of assigning a significant share of the arrivals to events when their are about to end (around 21:00). The proposed non-linear model (BAM-GP) makes a much more reasonable estimate with that respect. 
This is a consequence of the fact that the relation between arrivals to an event and its end time, which is used as one of the model features, is non-linear. Moreover, it is expected that this relation would be also dependent on the type of event, which is something that a simple linear model cannot capture. 

Finally, Fig.~\ref{fig:decomposition_bam} shows six additional illustrative decompositions produced by BAM-GP. All these examples further support the idea that BAM-GP produces reasonable and well-informed disaggregations of the total observed arrivals into the contributions of routine commuting and the effects of the various events. For example, Fig.~\ref{fig:marginals3} shows the case where BAM-GP estimates a very small localized contribution from the event, which is not surprising because this was a small-sized event with a very narrow target audience that took place in a not so popular venue. Similarly, in Figs.~\ref{fig:marginals4} and \ref{fig:marginals5}, the model assigns large shares to the ``Asia Pacific Food Expo" and ``Megatex", which is reasonable since the former is a particularly large food festival in Singapore and the latter is a popular electronics and IT showcase. Particularly, the difference between ``Megatex" and ``Living Tomorrow 2013" is quite significant, which is not surprising considering the popularity of the event and how much Singaporeans like technology. 

\section{Conclusions}

We proposed BAM-GP: a Bayesian additive model (BAM) with Gaussian process (GP) components that allows for an observed variable to be modeled as a sum of a variable number of non-linear functions on subsets of the input variables. We developed an efficient approximate inference algorithm using expectation propagation (EP), which allows us to both make predictions about the unobserved totals and to estimate the marginal distributions of the additive components. The proposed model is then capable of being flexible, while retaining its interpretability characteristics. We apply BAM-GP to the problem of estimating public transport arrivals in special event scenarios. Using a five months dataset of Singapore's ``tap-in/tap-out" fare card system and data about special events mined from the Web, we show that the model presented not only outperforms others that do not account for information about events, thus verifying the value of Internet-mined data for understanding urban mobility, but also outperforms other more general models that do account for events information. Furthermore, due to its additive nature and Bayesian formulation, BAM-GP is capable of estimating the posterior marginal distributions that correspond to routine commuting and the contributions of the various events, which is of great value for both public transport operators/planners and event organizers. Moreover, thanks to its flexibility, the proposed additive framework allows to incorporate domain knowledge constraints over the component values, such as non-negativity. Finally, we believe that the presented methodology is quite general and that it can be easily adapted beyond the transportation area such as, for example, in the analysis of financial time-series, cell-phone call records or electrical consumption signals. 

\appendices

\section{EP algorithm details}
\label{app:ep_nonlinear}

Here we provide additional details on the message-passing algorithm depicted in Fig.~\ref{fig:factor_graph_gp} for performing approximate inference in the proposed Bayesian additive model using EP. 

The first step consists on computing message from the $g^r$ and $g^e$ factors to the $f_n^r$ and $f_n^e$ variables respectively
\begin{align}
\label{eq:message1}
m_{g^r \rightarrow f_n^r}^t(f_n^r) &= \int p(\textbf{f}^r|\textbf{X}) \prod_{j \neq n} m_{f_j^r \rightarrow g^r}^{t-1}(f_j^r) \, df_j^r.
\end{align}
Conceptually, the combination of the prior $p(\textbf{f}^r|\textbf{X}) = \N(\textbf{f}^r|\textbf{0},\textbf{K}^r)$ and the $n-1$ messages in (\ref{eq:message1}) can be viewed in two ways, either by explicitly multiplying out the factors or, equivalently, by removing the $n^\mathrm{th}$ message from the approximate posterior on $\textbf{f}^r$ which, in general, is easier to compute. The approximate posterior on $\textbf{f}^r$ is given by
\begin{align}
\label{eq:approx_posterior}
q(\textbf{f}^r) &= \frac{1}{Z_{EP}} \N(\textbf{f}^r|\textbf{0},\textbf{K}^r) \prod_{n=1}^N \N\Big(f_n^r\Big|\mu_{f_n^r \rightarrow g^r}^{t-1},v_{f_n^r \rightarrow g^r}^{t-1}\Big) \nonumber\\
&= \N(\textbf{f}^r|\bs\mu^r,\bs\Sigma^r),\nonumber
\end{align}
with $\bs\mu^r = \bs\Sigma^r \big(\tilde{\bs\Sigma}^r\big)^{-1} \tilde{\bs\mu}^r$ and $\bs\Sigma^r = \big(\big(\textbf{K}^r\big)^{-1} + \big(\tilde{\bs\Sigma}^r\big)^{-1}\big)^{-1}$, 
where $\tilde{\bs\mu}^r$ is the vector of $\mu_{f_n^r \rightarrow g^r}^{t-1}$ and $\tilde{\bs\Sigma}^r$ is a diagonal matrix with $\tilde{\Sigma}_{nn}^r = v_{f_n^r \rightarrow g^r}^{t-1}$. Hence, the marginal for $f_n^r$ from $q(\textbf{f}^r)$ is given by $q(f_n^r) = \N\big(f_n^r\big|\mu_n^r,\Sigma_{nn}^r\big)$.
The message from the factor $g^r(\textbf{f}^r)$ to the $f_n^r$ variables is then given by
\begin{align}
m_{g^r \rightarrow f_n^r}^t(f_n^r) &= \frac{q(f_n^r)}{m_{f_n^r \rightarrow g^r}^{t-1}(f_n^r)},\nonumber
\end{align}
which can be easily computed by making use of the standard result for the division of two Gaussian distributions. Indeed, the same result can be applied to compute all the messages from variables to factors (see (\ref{eq:m_x_f})).

The remaining messages from factors to variables are of the form (or similar)
\begin{align}
m_{k_n \rightarrow y_n^{e_i}}^t(y_n^{e_i}) &= \int \N(y_n | y_n^r + \textstyle\sum\nolimits_{i=1}^{E_n} y_n^{e_i}, v) \, m_{y_n^r \rightarrow k_n}^t(y_n^r) \nonumber\\
&\times \prod_{j \neq i} m_{y_n^{e_j} \rightarrow k_n}^t(y_n^{e_j}) \, dy_n^r \, d\{y_n^{e_j}\}_{j \neq i},\nonumber\\
m_{h_n^r \rightarrow y_n^r}^t(y_n^r) &= \int p(y_n^r|f_n^r) \, m_{f_n^r \rightarrow h_n^r}^t(f_n^r) \, df_n^r.\nonumber
\end{align}
The former can be readily evaluated by noticing that
$\N(y_n | y_n^r + \textstyle\sum\nolimits_{i=1}^{E_n} y_n^{e_i}, v) = \N(y_n^{e_i} | y_n - y_n^r - \textstyle\sum\nolimits_{j \neq i} y_n^{e_j}, v)$, 
while the latter in general cannot be solved analytically, due to the non-Gaussian term $p(y_n^r|f_n^r)$. Fortunately, these messages do not need to be evaluated exactly. Instead, EP only requires that we are able to compute moments involving these messages (the $\mbox{proj}[\cdot]$ operator), which can be determined analytically as in (\ref{eq:moments_truncated}) for the truncated Gaussian, or by making use of numerical integration procedures in the Poisson case. For example, the approximate marginal distribution on $y_n^r$ is given by
\begin{align}
q^t(y_n^r) &= \mbox{proj}\bigg[ \mathbb{I}(y_n^r > 0) \, m_{k_n \rightarrow y_n^r}^{t-1}(y_n^r) \nonumber\\
&\times \int \N(y_n^r | f_n^r, \beta_r) \, m_{f_n^r \rightarrow h_n^r}^t(f_n^r) \, df_n^r  \bigg]\nonumber
\end{align}
which can be computed by first making use of the standard results for the marginal and product of two Gaussian, and then determining the moments of the resulting Gaussian distribution using (\ref{eq:moments_truncated}).

\ifCLASSOPTIONcompsoc
  \section*{Acknowledgments}
\else
  \section*{Acknowledgment}
\fi

{The Funda\c{c}\~ao para a Ci\^encia e Tecnologia (FCT) is gratefully acknowledged for founding this work with the grants SFRH/BD/78396/2011 and PTDC/ECM-TRA/1898/2012 (InfoCROWDS).}


\ifCLASSOPTIONcaptionsoff
  \newpage
\fi



\bibliographystyle{IEEEtran}
\bibliography{bam_final.bbl}

\begin{thebibliography}{10}
\providecommand{\url}[1]{#1}
\csname url@samestyle\endcsname
\providecommand{\newblock}{\relax}
\providecommand{\bibinfo}[2]{#2}
\providecommand{\BIBentrySTDinterwordspacing}{\spaceskip=0pt\relax}
\providecommand{\BIBentryALTinterwordstretchfactor}{4}
\providecommand{\BIBentryALTinterwordspacing}{\spaceskip=\fontdimen2\font plus
\BIBentryALTinterwordstretchfactor\fontdimen3\font minus
  \fontdimen4\font\relax}
\providecommand{\BIBforeignlanguage}[2]{{%
\expandafter\ifx\csname l@#1\endcsname\relax
\typeout{** WARNING: IEEEtran.bst: No hyphenation pattern has been}%
\typeout{** loaded for the language `#1'. Using the pattern for}%
\typeout{** the default language instead.}%
\else
\language=\csname l@#1\endcsname
\fi
#2}}
\providecommand{\BIBdecl}{\relax}
\BIBdecl

\bibitem{krygsman2004}
S.~Krygsman, M.~Dijst, and T.~Arentze, ``Multimodal public transport: an
  analysis of travel time elements and the interconnectivity ratio,''
  \emph{Transport Policy}, vol.~11, no.~3, pp. 265--275, 2004.

\bibitem{kwon2006}
J.~Kwon, M.~Mauch, and P.~Varaiya, ``Components of congestion: delay from
  incidents, special events, lane closures, weather, potential ramp metering
  gain, and excess demand,'' \emph{Transportation Research Record}, vol. 1959,
  no.~1, pp. 84--91, 2006.

\bibitem{pereira2013JITS}
F.~Pereira, F.~Rodrigues, and M.~Ben-Akiva, ``Using data from the web to
  predict public transport arrivals under special events scenarios,'' \emph{J.
  of Intelligent Transportation Systems}, 2013.

\bibitem{Bishop2013}
C.~Bishop, ``Model-based machine learning,'' \emph{Philosophical Transactions
  of the Royal Society A: Mathematical, Physical \& Engineering Sciences}, vol.
  371, no. 1984, 2013.

\bibitem{Minka2001}
T.~P. Minka, ``Expectation propagation for approximate bayesian inference,'' in
  \emph{Proc. of the 17th Conference in Uncertainty in Artificial
  Intelligence}, ser. UAI '01.\hskip 1em plus 0.5em minus 0.4em\relax Morgan
  Kaufmann, 2001, pp. 362--369.

\bibitem{Rasmussen2005}
C.~E. Rasmussen and C.~Williams, \emph{{Gaussian processes for machine
  learning}}.\hskip 1em plus 0.5em minus 0.4em\relax The MIT Press, 2005.

\bibitem{ben-akiva-lerman-1987}
M.~Ben-Akiva and S.~Lerman, \emph{Discrete Choice Analysis: Theory and
  Application to Travel Demand}.\hskip 1em plus 0.5em minus 0.4em\relax The MIT
  Press, 1987.

\bibitem{Kuppam2011}
A.~Kuppam, R.~Copperman, T.~Rossi, V.~Livshits, L.~Vallabhaneni, T.~Brown, and
  K.~DeBoer, ``Innovative methods for collecting data and for modeling travel
  related to special events,'' \emph{Transportation Research Record}, vol.
  2246, no.~1, pp. 24--31, 2011.

\bibitem{chang2013}
M.~Chang and P.~Lu, ``A multinomial logit model of mode and arrival time
  choices for planned special events,'' \emph{J. of the Eastern Asia Society
  for Transportation Studies}, vol.~10, pp. 710--727, 2013.

\bibitem{CalabresePervasive2010}
F.~Calabrese, F.~Pereira, G.~D. Lorenzo, L.~Liu, and C.~Ratti, ``{The geography
  of taste: analyzing cell-phone mobility and social events},'' in
  \emph{Pervasive Computing}.\hskip 1em plus 0.5em minus 0.4em\relax Springer,
  2010, pp. 22--37.

\bibitem{FHWA06}
FHWA, ``Planned special events: Checklists for practitioners,'' Federal Highway
  Administration, Tech. Rep. FHWA-HOP-06-113, 2006.

\bibitem{CoutroubasEtAl03}
F.~Coutroubas, G.~Karabalasis, and Y.~Voukas, ``Public transport planning for
  the greatest event - the 2004 olympic games,'' in \emph{European Transp.
  Conference}, 2003.

\bibitem{PotierEtAl03}
F.~Potier, P.~Bovy, and C.~Liaudat, ``Big events: planning, mobility
  management,'' in \emph{European Transp. Conference}, 2003.

\bibitem{Bolte2006}
C.~Levecq, B.~Kuhn, and D.~Jasek, ``General guidelines for active traffic
  management deployment,'' Texas Transportation Institute, Tech. Rep. UTCM
  10-01-54-1, 2011.

\bibitem{Middleham2006}
P.~Brinckerhoff, ``Synthesis of active traffic management experiences in europe
  and the united states,'' Federal Highway Administration, Tech. Rep.
  FHWA-HOP-10-031, 2010.

\bibitem{pereira2015so}
F.~Pereira, F.~Rodrigues, E.~Polisciuc, and M.~Ben-Akiva, ``Why so many people?
  explaining nonhabitual transport overcrowding with internet data,''
  \emph{IEEE Transactions on Intelligent Transportation Systems}, vol.~16,
  no.~3, pp. 1370--1379, 2015.

\bibitem{InferNET12}
T.~Minka, J.~Winn, J.~Guiver, and D.~Knowles, ``{Infer.NET 2.5},'' 2012,
  microsoft Research Cambridge. http://research.microsoft.com/infernet.

\bibitem{Hastie1990}
T.~Hastie and R.~Tibshirani, \emph{Generalized additive models}.\hskip 1em plus
  0.5em minus 0.4em\relax CRC Press, 1990, vol.~43.

\bibitem{Hastie2003}
T.~Hastie, R.~Tibshirani, and J.~Friedman, \emph{{The Elements of Statistical
  Learning: Data Mining, Inference, and Prediction}}.\hskip 1em plus 0.5em
  minus 0.4em\relax Springer, 2003.

\bibitem{Ravikumar2009}
P.~Ravikumar, J.~Lafferty, H.~Liu, and L.~Wasserman, ``Sparse additive
  models,'' \emph{Journal of the Royal Statistical Society: Series B
  (Statistical Methodology)}, vol.~71, no.~5, pp. 1009--1030, 2009.

\bibitem{Duvenaud2011}
D.~Duvenaud, H.~Nickisch, and C.~Rasmussen, ``Additive gaussian processes,'' in
  \emph{Advances in neural information processing systems}, 2011, pp. 226--234.

\bibitem{Chipman2010}
H.~Chipman, E.~George, and R.~McCulloch, ``Bart: Bayesian additive regression
  trees,'' \emph{The Annals of Applied Statistics}, pp. 266--298, 2010.

\bibitem{zheng2015methodologies}
Y.~Zheng, ``Methodologies for cross-domain data fusion: an overview,''
  \emph{IEEE Trans. on Big Data}, vol.~1, no.~1, pp. 16--34, 2015.

\bibitem{zheng2014urban}
Y.~Zheng, L.~Capra, O.~Wolfson, and H.~Yang, ``Urban computing: concepts,
  methodologies, and applications,'' \emph{ACM Transactions on Intelligent
  Systems and Technology (TIST)}, vol.~5, no.~3, p.~38, 2014.

\bibitem{zheng2013u}
Y.~Zheng, F.~Liu, and H.-P. Hsieh, ``U-air: When urban air quality inference
  meets big data,'' in \emph{ACM SIGKDD Int. Conf. on Knowledge discovery and
  data mining}.\hskip 1em plus 0.5em minus 0.4em\relax ACM, 2013, pp.
  1436--1444.

\bibitem{Van2015}
N.~van Oort, T.~Brands, and E.~de~Romph, ``Short term ridership prediction in
  public transport by processing smart card data,'' in \emph{Transportation
  Research Board Annual Meeting}, no. 15-0773, 2015.

\bibitem{zhong2014signal}
M.~Zhong, N.~Goddard, and C.~Sutton, ``Signal aggregate constraints in additive
  factorial hmms, with application to energy disaggregation,'' in
  \emph{Advances in Neural Information Processing Systems}, 2014, pp.
  3590--3598.

\bibitem{Barber2012}
D.~Barber, \emph{{Bayesian reasoning and machine learning}}.\hskip 1em plus
  0.5em minus 0.4em\relax Cambridge University Press, 2012.

\bibitem{ide2009travel}
T.~Id{\'e} and S.~Kato, ``Travel-time prediction using gaussian process
  regression: A trajectory-based approach.'' in \emph{SDM}.\hskip 1em plus
  0.5em minus 0.4em\relax SIAM, 2009, pp. 1185--1196.

\bibitem{xie2010gaussian}
Y.~Xie, K.~Zhao, Y.~Sun, and D.~Chen, ``Gaussian processes for short-term
  traffic volume forecasting,'' \emph{Transportation Research Record}, no.
  2165, pp. 69--78, 2010.

\bibitem{neumann2009stacked}
M.~Neumann, K.~Kersting, Z.~Xu, and D.~Schulz, ``Stacked gaussian process
  learning,'' in \emph{Data Mining, 2009. ICDM'09. Ninth IEEE International
  Conference on}.\hskip 1em plus 0.5em minus 0.4em\relax IEEE, 2009, pp.
  387--396.

\bibitem{Murphy2012}
K.~P. Murphy, \emph{Machine Learning: A Probabilistic Perspective}.\hskip 1em
  plus 0.5em minus 0.4em\relax The MIT Press, 2012.

\bibitem{winkler1990string}
W.~E. Winkler, ``String comparator metrics and enhanced decision rules in the
  fellegi-sunter model of record linkage,'' in \emph{Proc. of the Section on
  Survey Research Methods}.\hskip 1em plus 0.5em minus 0.4em\relax ERIC, 1990,
  pp. 354--359.

\bibitem{Blei:2003:LDA:944919.944937}
D.~M. Blei, A.~Y. Ng, and M.~I. Jordan, ``Latent dirichlet allocation,''
  \emph{J. Mach. Learn. Res.}, vol.~3, pp. 993--1022, 2003.

\bibitem{Borg2005}
I.~Borg and P.~Groenen, \emph{Modern multidimensional scaling: Theory and
  applications}.\hskip 1em plus 0.5em minus 0.4em\relax Springer Science \&
  Business Media, 2005.

\end{thebibliography}
%
%
%

%

\begin{IEEEbiography}[{\includegraphics[width=1in,height=1.25in,clip,keepaspectratio]{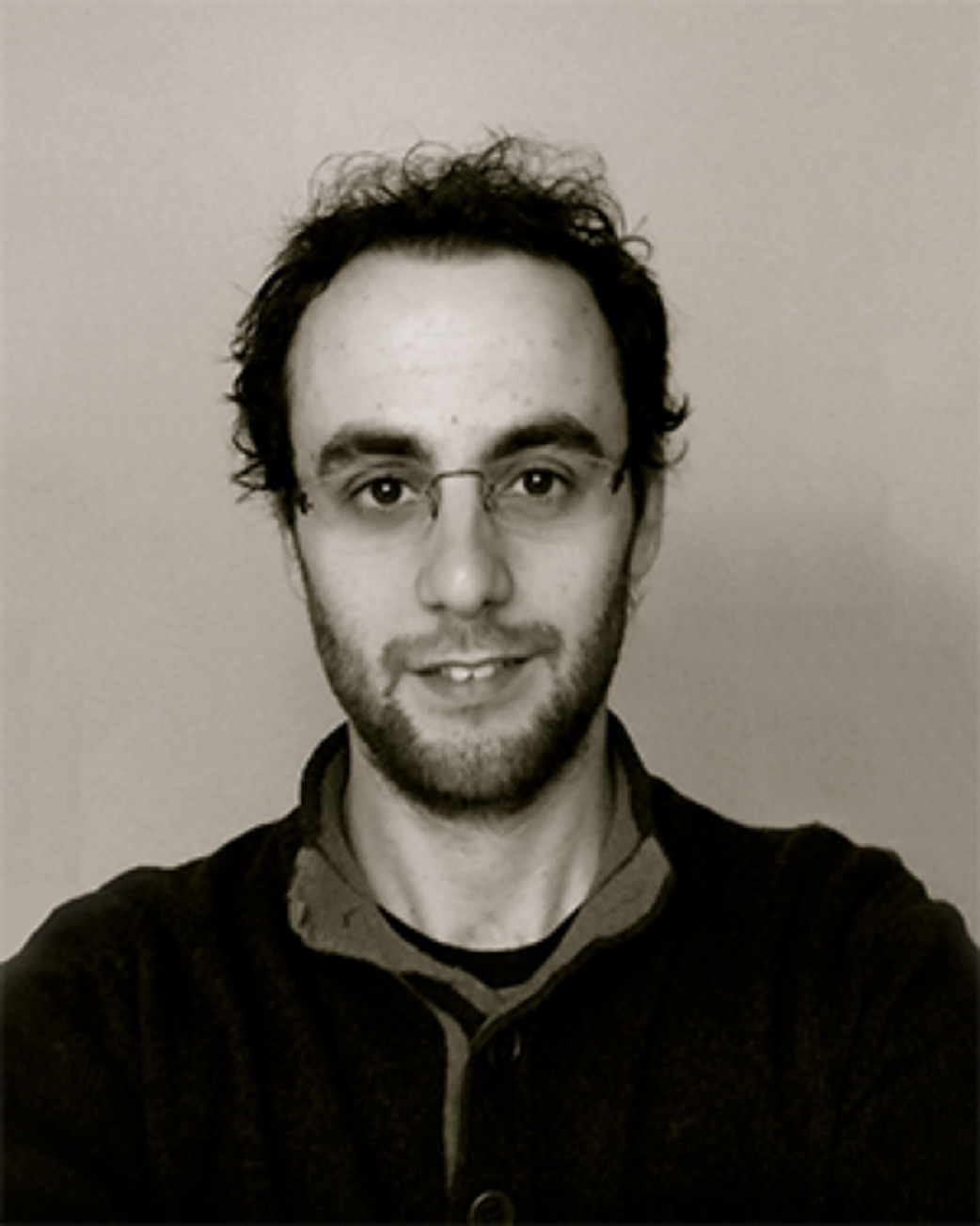}}]{Filipe~Rodrigues}
is Postdoctoral Fellow at Technical University of Denmark, where he is working on machine learning models for understanding urban mobility and the behaviour of crowds, with emphasis on the effect of special events in mobility and transportation systems. He received a Ph.D. degree in Information Science and Technology from University of Coimbra, Portugal, where he developed probabilistic models for learning from crowdsourced and noisy data. His research interests include machine learning, probabilistic graphical models, natural language processing, intelligent transportation systems and urban mobility. 
\end{IEEEbiography}

\begin{IEEEbiography}[{\includegraphics[width=1in,height=1.25in,clip,keepaspectratio]{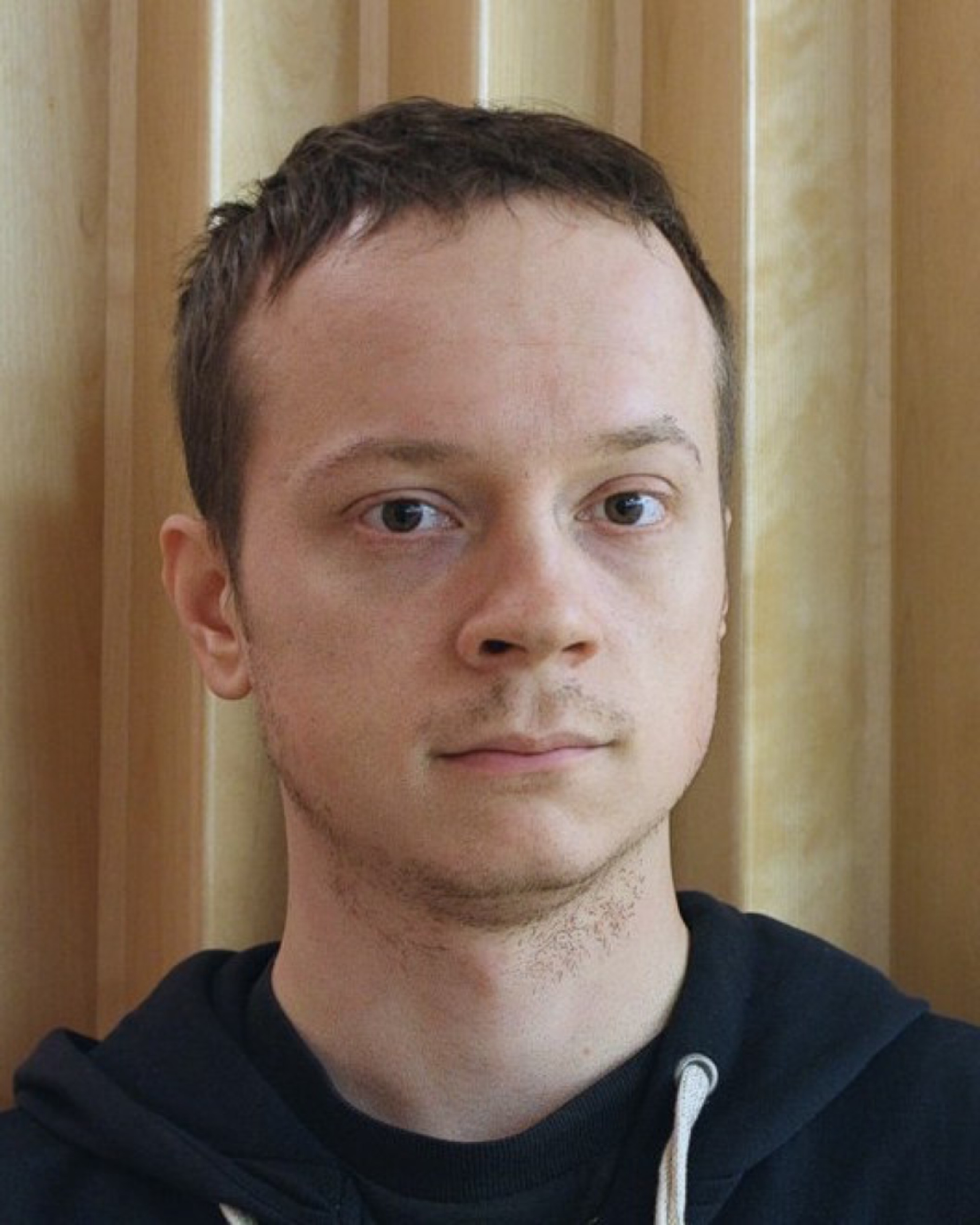}}]{Stanislav~S.~Borysov} is Postdoctoral Fellow at the Nordic institute for theoretical physics NORDITA (Stockholm, Sweden) where he is working on data mining, statistical learning and optimization tools for complex physical, biological and social systems. Previously, he also was a Postdoctoral Associate at the Singapore-MIT Alliance for Research and Technology, where he was working on data driven and statistical approaches for intelligent transportation systems. He received a Ph.D. degree in Theoretical Physics from the Sumy State University and Institute of Applied Physics, National Academy of Science of Ukraine (Sumy, Ukraine) for studies related to statistical physics of complex and nonlinear systems.
\end{IEEEbiography}

\begin{IEEEbiography}[{\includegraphics[width=1in,height=1.25in,clip,keepaspectratio]{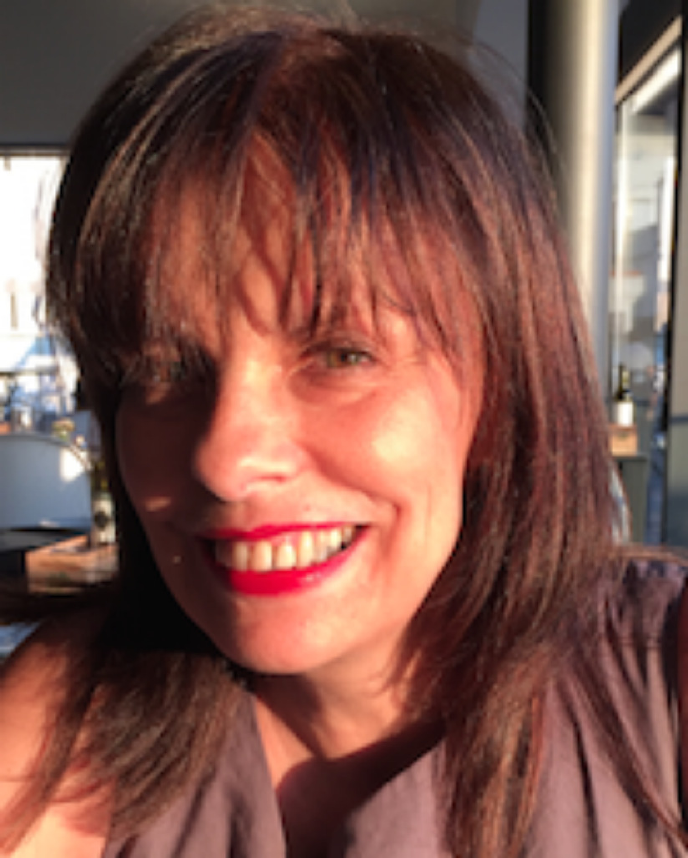}}]{Bernardete~Ribeiro}
is Associate Professor at the Informatics Engineering Department, University of Coimbra in Portugal, from where she received a D.Sc. in Informatics Engineering, a Ph.D. in Electrical Engineering, speciality of Informatics, and a MSc in Computer Science. Her research interests are in the areas of Machine Learning, Pattern Recognition and Signal Processing and their applications to a broad range of fields. She was responsible/participated in several research projects in a wide range of application areas such as Text Classification, Financial, Biomedical and Bioinformatics. Bernardete Ribeiro is IEEE Senior Member, and member of IARP International Association of Pattern Recognition and ACM.
\end{IEEEbiography}


\begin{IEEEbiography}[{\includegraphics[width=1in,height=1.25in,clip,keepaspectratio]{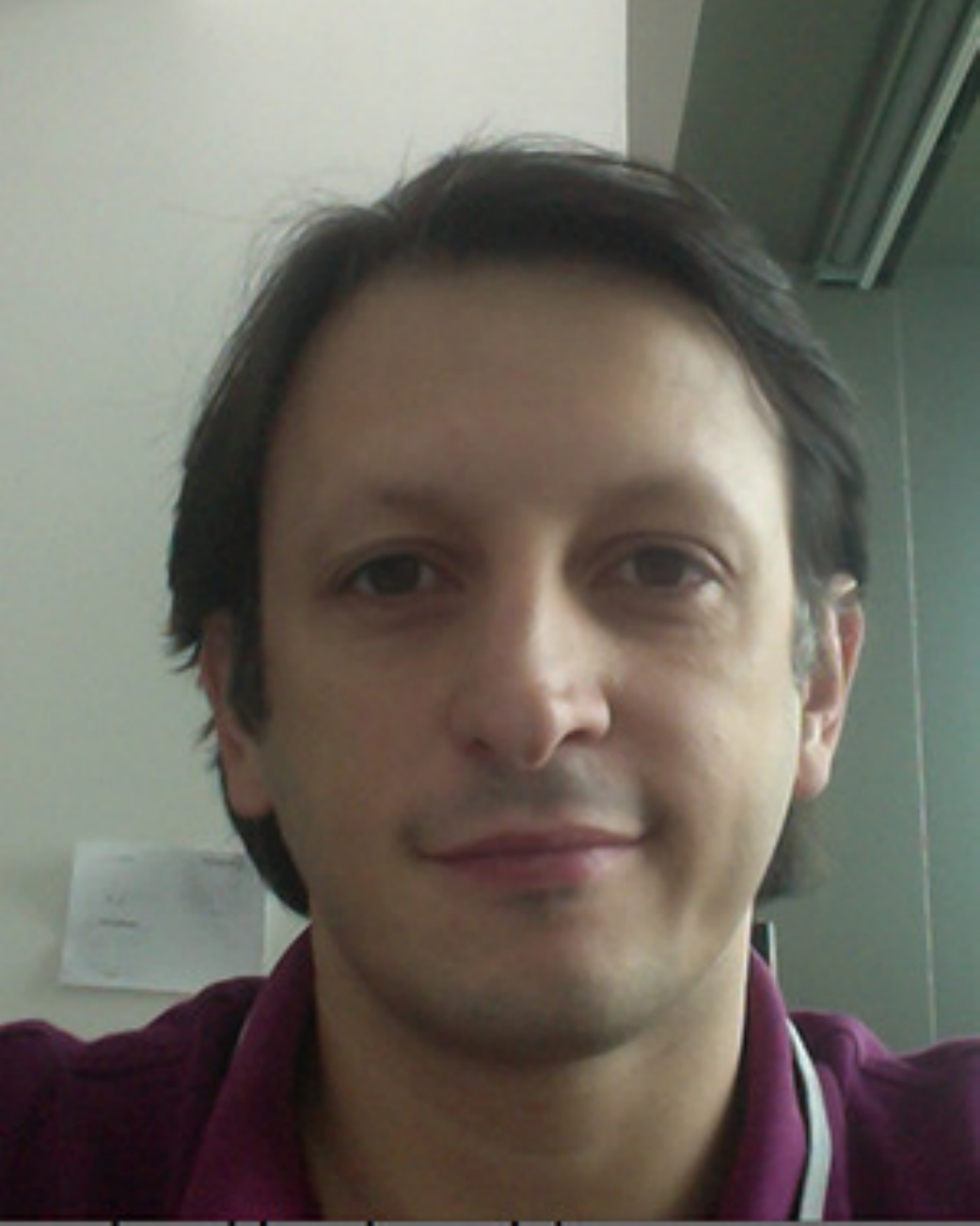}}]{Francisco~C.~Pereira}
is Full Professor at the Technical University of Denmark (DTU), where he leads the Smart Mobility research group. His main research focus is on applying machine learning and pattern recognition to the context of transportation systems with the purpose of understanding and predicting mobility behavior, and modeling and optimizing the transportation system as a whole. He has Master’s (2000) and Ph.D. (2005) degrees in Computer Science from University of Coimbra, and has authored/co-authored over 70 journal and conference papers in areas such as pattern recognition, transportation, knowledge based systems and cognitive science. Francisco was previously Research Scientist at MIT and Assistant Professor in University of Coimbra. He was awarded several prestigious prizes, including an IEEE Achievements award, in 2009, the Singapore GYSS Challenge in 2013, and the Pyke Johnson award from Transportation Research Board, in 2015.
\end{IEEEbiography}




\end{document}